%% file: main_neurips.tex
\documentclass{article}

\usepackage[preprint]{neurips_2026}

\usepackage[utf8]{inputenc} 
\usepackage[T1]{fontenc}    
\usepackage{hyperref}       
\usepackage{url}            
\usepackage{booktabs}       
\usepackage{amsfonts}       
\usepackage{nicefrac}       
\usepackage{microtype}      
\usepackage{xcolor}         

\usepackage{graphicx}
\usepackage{subcaption}
\usepackage{comment}
\usepackage{CJKutf8}
\usepackage{multirow}
\usepackage[T1]{fontenc}
\usepackage{placeins}
\usepackage{tabularx}

\usepackage{amsmath}
\usepackage{amssymb}
\usepackage{mathtools}
\usepackage{amsthm}

\usepackage[textsize=tiny]{todonotes}

\newcommand{\LLM}{\textup{LLM}}

\newcommand{\prompt}{\textit{prompt}}
\newcommand{\trajectories}{\textit{trajectories}}
\newcommand{\feedback}{\textit{feedback}}
\newcommand{\Sharded}{\texttt{Sharded}}

\newcommand{\MC}{\textup{MC}}
\newcommand{\TD}{\textup{TD}}

\usepackage{algorithm}
\usepackage{algorithmic}
\title{Prompt Reinforcing for Long-term Planning of Large Language Models}

\author{%
  Hsien-Chin Lin\dag\thanks{\texttt{linh@hhu.de}}, 
  Nurul Lubis\dag,  
Benjamin Matthias Ruppik\dag, 
Carel van Niekerk\dag, \\
\textbf{Chia-Hao Shen\ddag, 
Michael Heck\dag, 
Renato Vukovic\dag, 
Shutong Feng\dag,  
Milica Ga\v{s}i\'{c}\dag} \\
\dag Heinrich-Heine-Universit\"{a}t D\"{u}sseldorf, \ddag Independent researcher \\
}

\begin{document}

\maketitle

\begin{abstract}
  Large language models (LLMs) have achieved remarkable success in a wide range of natural language processing tasks and can be adapted through prompting. 
However, they remain suboptimal in multi-turn interactions, often relying on incorrect early assumptions and failing to track user goals over time, which makes such tasks particularly challenging.
Prior works in dialogue systems have shown that long-term planning is essential for handling interactive tasks. 
In this work, we propose a prompt optimisation framework inspired by reinforcement learning, which enables such planning to take place by only modifying the task instruction prompt of the LLM-based agent. 
By generating turn-by-turn feedback and leveraging experience replay for prompt rewriting, our proposed method shows significant improvement in multi-turn tasks such as text-to-SQL and task-oriented dialogue. 
Moreover, it generalises across different LLM-based agents and can leverage diverse LLMs as meta-prompting agents.
This warrants future research in reinforcement learning-inspired parameter-free optimisation methods.
\end{abstract}

\input{tex/0_introduction}

\input{tex/1_related_work}
\input{tex/2_proposed_method}
\input{tex/3_exp-setting}
\input{tex/4_results}
\input{tex/5_conclusion}
\bibliography{reference}
\bibliographystyle{apalike}
\appendix
\input{tex/appendix/F_limitations}

\input{tex/appendix/A_model_list}
\input{tex/appendix/D_RPO_algorithm}

\input{tex/appendix/B_converge}
\input{tex/appendix/E_qualitative}
\input{tex/appendix/C_prompts}



\end{document}

%% file: tex/0_introduction.tex
\section{Introduction}
\label{sec:intro}
Large language models (LLMs) have shown an extraordinary ability to perform a wide range of tasks, from generating images in various styles to writing code in different programming languages for diverse purposes. 
LLMs are typically post-trained using reinforcement learning from human feedback (RLHF) \citep{ouyang2022training}, where they receive single-turn rewards for individual responses rather than rewards reflecting the quality of an entire multi-turn conversation.
This limits their effectiveness in interactions where tasks are underspecified and clarified over time, often leading to early mistakes, incorrect assumptions, and cascading failures~\citep{laban2025llmslostmultiturnconversation}.
On the other hand, prior work in dialogue systems demonstrates that long-term planning is vital for interactive tasks, making it essential for LLMs \citep{young2002talking,6407655}.

Directly optimising model parameters can improve long-horizon behaviour (e.g., LoRA, DPO, and dialogue-level RL) \citep{hu2022lora,feng2025empoweringllmstaskorienteddialogues,feng2025emotionallyintelligenttaskorienteddialogue}, but it is often too expensive for real-time updates and infeasible for API-only models.

Gradient-free self-feedback methods avoid weight updates \citep{peng2023check, NEURIPS2023_1b44b878, yao2023react, elizabeth2025exploring}, but often require repeated feedback loops during inference.
Existing prompt-optimisation methods mostly target input-output tasks and do not explicitly address long-horizon multi-turn planning \citep{yang2024large, tang2024unleashing, pryzant-etal-2023-automatic, yuksekgonul2025optimizing}.

\begin{figure*}[h]
    \centering
    \includegraphics[width=0.7\linewidth]{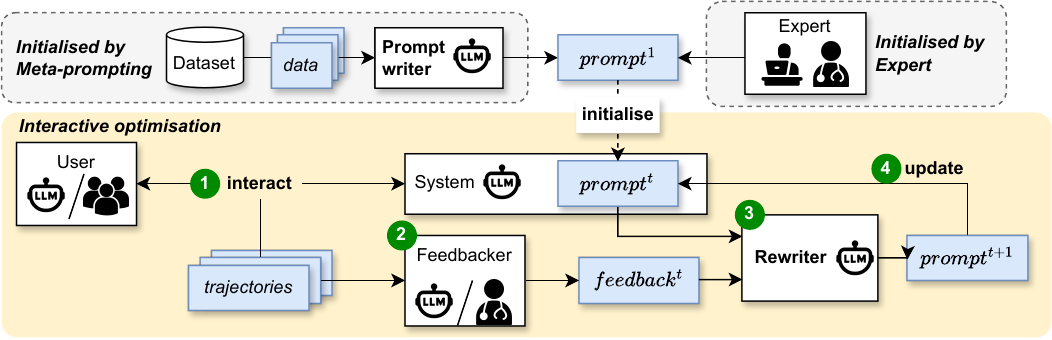}
    \caption{
        \textbf{Structure of Reinforced Prompt Optimisation (RPO).}
        The initial $\prompt^1$ can be written by LLMs or experts.
        In interactive optimisation, the system will first interact with the environment (simulated or real users). 
        Then, the feedbacker (human or LLM) will provide textual feedback based on trajectories, and the rewriter generates a new prompt based on the original prompt and the textual feedback to update the system's original prompt.
        One cycle of interactive optimisation is called an epoch, denoted by \emph{superscripts}.
    }
    \label{fig:TaP-structure}
\end{figure*}

To address these limitations, we propose Reinforced Prompt Optimisation (RPO), illustrated in \autoref{fig:TaP-structure}.
RPO improves long-horizon behaviour by iteratively refining the task instruction from natural-language feedback, starting from either expert-written or automatically generated prompts~\citep{zhou2023large, pryzant-etal-2023-automatic, ye-etal-2024-prompt}.

In RPO, an LLM-based system interacts with an environment (real or simulated users).
A feedbacker (human or LLM) provides turn-level textual feedback inspired by temporal difference (TD).
As shown in the right part of \autoref{fig:feedback-pipeline}, for each turn $t_i$, the LLM-generated feedback includes: 
(1) predicted user emotion in the next turn elicited by the system response $a_i$, (2) a forecast of dialogue success or failure, and (3) suggestions based on the subdialogue $t_{1:i}$. 
These are then aggregated into dialogue-level feedback.

A separate LLM-based rewriter refines the prompt based on the feedback and the previous prompt. 
Experience replay is applied by leveraging feedback–prompt pairs from both the current and past iterations.
The updated prompt is used in future interactions (Section~\ref{sec:proposed-method-detail}).
Inspired by these well-studied reinforcement learning concepts, the goal of RPO is to effectively strengthen the system agent’s long-term planning ability and overall task success.

Our contributions are as follows:
\begin{itemize}
    \item We propose Reinforced Prompt Optimisation (RPO), a meta-prompting framework that improves multi-turn planning by iteratively updating task instructions from textual feedback.
    \item We study two design choices for feedback-driven prompt updates: turn-level TD-style feedback and replay-based rewriting.
    \item RPO works with external expert signals and generalises across both system backbones and meta-prompting LLMs, including API-only settings.
\end{itemize}

%% file: tex/1_related_work.tex
\section{Related work}
\label{sec:related-work}

\paragraph{Gradient-based optimisation for LLMs}
Full fine-tuning of large models is expensive, so parameter-efficient methods (e.g., adapters and LoRA) are widely used \citep{hu2022lora, hu-etal-2023-llm, lialin2023scaling}. 
Continuous prompting methods such as prefix-tuning and soft prompts can also approach fine-tuning performance \citep{lester-etal-2021-power,qin-eisner-2021-learning,li-liang-2021-prefix, LIU2023,liu-etal-2022-p}.
However, these methods are not applicable to API-only models and are hard to deploy for rapid online updates.

\paragraph{Self-feedback}
To improve the performance of text-based prompts, various prompting styles are proposed, e.g., Chain-of-Thought \citep{NEURIPS2022_9d560961} or ReAct \citep{yao2023react}. 
These prompting methods encourage LLMs to reason before taking action or generating responses, which leads to better performance. 
However, optimising the prompt for better performance by manual trial and error is inefficient.
Instead, self-feedback methods are introduced to refine the LLMs' response, e.g.,
LLM-augmenter generates feedback by itself and leverages external knowledge to rewrite its response \citep{peng2023check}, and Reflexion
summarises previous interactions with the environment as `reflections' to improve the model's response \citep{NEURIPS2023_1b44b878, NEURIPS2023_91edff07}. 

While this demonstrates the ability of LLMs for self-correction, these self-feedback methods rely on frequent API calls since their original prompt is not optimal. 
As a result, the computation cost and latency during inference are not negligible.

\paragraph{Prompt optimisation}
Meta-prompting methods are widely used to generate a prompt without human editing. 
The automatic prompt engineer (APE) method leverages an LLM, which is instructed to generate an initial prompt and selects the prompt with the best performance on the target task \citep{zhou2023large}. 
Automatic prompt optimisation (APO) further employs a self-feedback module to provide textual feedback, which gives suggestions on how to edit the old prompt \citep{pryzant-etal-2023-automatic}.
\citet{ye-etal-2024-prompt} propose a meta-prompt LLM to edit the original prompt step-by-step.
\citet{kong-etal-2024-prewrite} and \citet{cheng-etal-2024-black} train a sequence-to-sequence model for prompt rewriting by reinforcement learning and preference data, respectively.
\citet{yang2024large} propose optimisation by prompting (OPRO), which leverages LLMs to rewrite the original prompt based on a corresponding performance score. 
To leverage experience, \citet{NEURIPS2023_f6b22ac3} model LLMs as semi-parametric RL agents with memory storing task data, actions, and $Q$-value estimates for few-shot in-context learning.
\citet{zhang2024agentpro} propose Agent-Pro, which constructs policy-level reflections according to the numerical feedback from the environment and improves its policy incrementally.
\citet{tang2024unleashing} introduce the Gradient-inspired LLM-based Prompt Optimizer (GPO), which updates the prompt iteratively based on numerical feedback and controls the edit distance through a cosine-based decay strategy. 
TextGrad generates textual feedback based on the user input and system output for prompt rewriting \citep{yuksekgonul2025optimizing}.
Although these methods demonstrate promising performance in generating or improving prompts, they focus on single-turn tasks.
Our approach addresses multi-turn interactions, where prompts are updated with temporally grounded feedback to enhance long-term planning ability.

\paragraph{Learning ability of LLMs via prompting}

Although transformers are universal approximators \citep{Yun2020Are} and in-context learning in LLMs can be viewed as implicit fine-tuning \citep{dai-etal-2023-gpt}, the following remain open questions:
Can we prompt LLMs for arbitrary tasks, and what are the limitations of in-context learning?

\citet{petrov2024when} highlight the limitations of context-based fine-tuning methods, e.g., in-context learning, prompting, and prefix tuning, for new task learning in transformers. 
Specifically, transformers struggle to acquire new tasks solely through prompting, as prompts cannot change the model's attention patterns. 
Instead, they can only bias the output of the attention layers in a fixed direction and elicit skills learned through pre-training. 
Namely, only models with billions of parameters trained on vast, diverse datasets are capable of in-context learning, adapting to new tasks through examples or instructions without modifying their underlying weights. 
Therefore, we focus on fundamental models large enough to demonstrate their in-context learning ability, to investigate reinforcement prompt optimisation, which is fully composed of in-context learning with LLMs.

%% file: tex/2_proposed_method.tex
\section{Reinforced Prompt Optimisation}
\label{sec:proposed-method-detail}

Inspired by the gradient-based optimisation and reinforcement learning algorithms, where a model is initialised from pretraining and then further updated by on-policy learning based on interactions with the environment, we propose the Reinforced Prompt Optimisation (RPO) method (as shown in \autoref{fig:TaP-structure} and the pseudo code can be found in Algorithm~\ref{alg:rpo}). 
The initial instruction can be generated by a prompt writer $\LLM_{P}$ such as the automatic prompt engineer (APE) \citep{zhou2023large} (the upper left part of \autoref{fig:TaP-structure}) or written by human experts (the upper right part of \autoref{fig:TaP-structure}).

In each optimisation epoch, the \textbf{system} interacts with an environment (human or simulated users) and produces multi-turn $\trajectories$.
The \textbf{feedbacker} ($\LLM_F$ or human expert) generates textual feedback, and the \textbf{rewriter} ($\LLM_R$) updates the prompt accordingly.

Unlike self-refine methods that edit model outputs, RPO edits the task instruction itself.
Treating the instruction as a textual parameter reduces serving overhead by avoiding repeated feedback loops at inference time.

\subsection{Feedback generation}
\label{sec:feedback-generation}

\begin{figure}[t]
    \centering
    \includegraphics[width=0.9\linewidth]{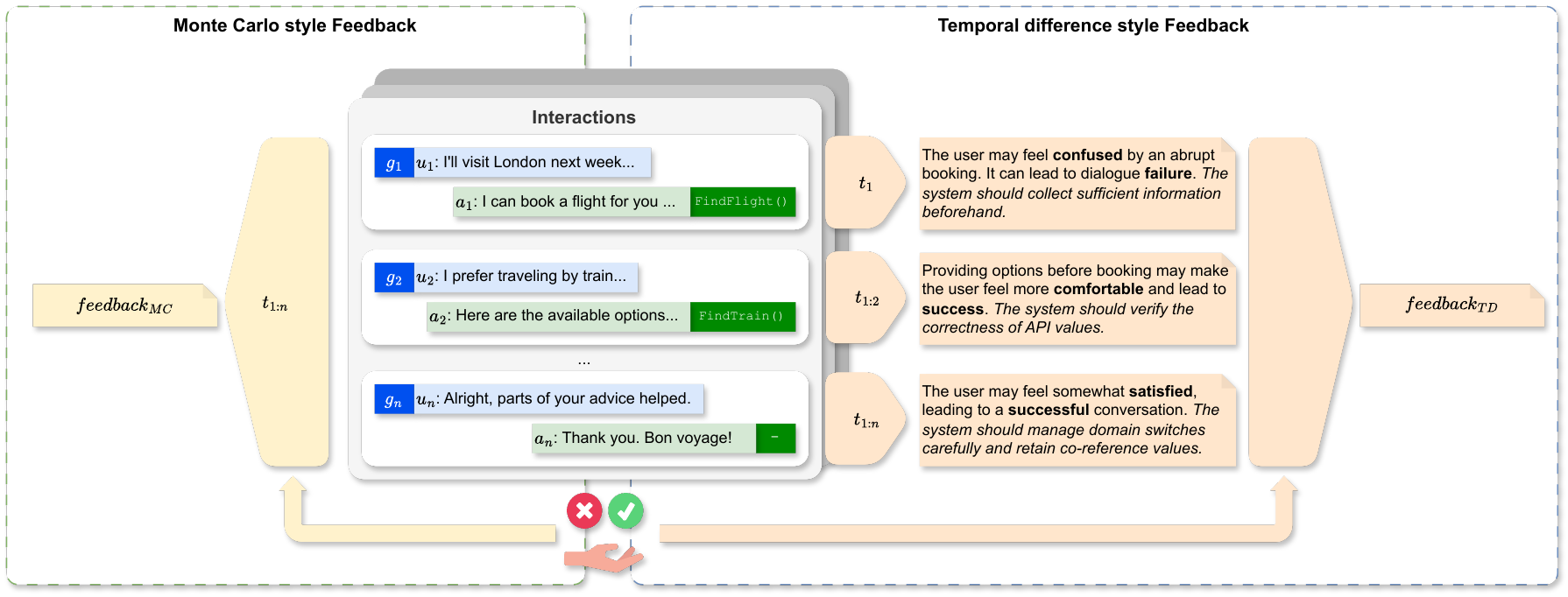}
    \caption{
        Workflow of feedback generation by an LLM. The Monte Carlo–style feedback (left) is generated after the entire interaction is completed, whereas the Temporal Difference–style feedback (right) consists of turn-level sub-feedback. Each turn includes natural-language utterances and, when available, the user's goal (in blue) and API calls (in green). Each sub-feedback includes a justification, two predictions (next-turn user satisfaction and goal success), and an actionable suggestion.
        \label{fig:feedback-pipeline}
    } 
\end{figure}

As shown in \autoref{fig:feedback-pipeline}, we consider two feedback schemes: Monte Carlo (MC)-style and Temporal Difference (TD)-style.

The \textbf{MC-style feedback} is produced only after the entire dialogue trajectory $(t_{1:n})$ has been completed (the prompt of the MC-style feedbacker is shown in \autoref{fig:mc-feedbacker}):
\begin{equation}
    \feedback_{\MC} = \LLM_F(t_{1:n})
\end{equation}
This approach is common in single-turn tasks \citep{pryzant-etal-2023-automatic,ye-etal-2024-prompt,wang2024promptagent,tang2024unleashing,yuksekgonul2025optimizing} and yields edits from a global success/failure signal.
It captures overall quality but collapses multi-turn interactions into a single outcome.

In contrast, our proposed \textbf{TD-style feedback} incorporates turn-level evaluations:
\begin{equation}
    \feedback_{\TD,j} = \LLM_F(t_1, \feedback_{\TD,1}, t_2, \feedback_{\TD,2}, \dots, t_j),
    \label{equ:TD_feedback}
\end{equation}

where $\feedback_{\TD,j}$ is the turn-level feedback at turn $j$.
All turn-level feedback, $\feedback_{\TD,1:j}$, is then summarised into final dialogue-level feedback $\feedback_{\TD}$ (prompt details in \autoref{fig:td-feedbacker}; examples in \autoref{fig:feedback-example}, \autoref{fig:feedback-example-dialogue}, and \autoref{fig:feedback-example-final}).
Instead of waiting for dialogue completion, TD-style feedback provides incremental sentiment/success predictions and actionable suggestions.

TD-style feedback can be interpreted as combining short-term and long-term signals \citep{ghazarian-etal-2022-wrong}:
\begin{equation}
\delta_t = r_t + \gamma V(s_{t+1}) - V(s_t)
\end{equation}
where $r_t$ is a short-term signal (e.g., user sentiment), $V(s_t)$ is the current value estimate, and $V(s_{t+1})$ is the estimated long-term value\footnote{For episodic tasks, $\gamma$ can be treated as $1$~\citep{sutton2018reinforcement}.}.

\subsection{Applying Feedback to the Prompt}

Unlike gradient-based optimisation, where gradients can be added or subtracted from model parameters, incorporating textual feedback into prompts is non-trivial. 
One cannot concatenate or remove arbitrary text from the original prompt without risking incoherence or loss of functionality.  
To address this, we introduce a basic \emph{rewriter} $\LLM_R$ to apply textual feedback on the original prompt:  
\begin{equation}
    \prompt^{i+1} = \LLM_R(\prompt^i, \feedback^i),
\end{equation}
where $i$ denotes the epoch index. 
Its instruction is shown in \autoref{fig:basic-rewriter}.

Inspired by experience \textbf{replay} in reinforcement learning~\citep{andrychowicz2017hindsight}, the rewriter can leverage not only the prompt and feedback from the current epoch, but also those from previous epochs (its instruction is shown in \autoref{fig:replay-rewriter}):  
\begin{equation}
\begin{split}
    \prompt^{i+1} 
    = 
    &\LLM_R(\prompt^i, \feedback^i, \prompt^{i-1}, \\
    &\feedback^{i-1}, \ldots, \prompt^1, \feedback^1).
\end{split}
\label{equ:replay-rewriter}
\end{equation}

RPO reduces task-specific manual prompt engineering by automating prompt refinement from environment feedback (simulated or human).
The feedbacker and rewriter prompts are task-agnostic and specified once; optimising those meta-prompts is left to future work.

%% file: tex/3_exp-setting.tex
\section{Experiment Settings}
\label{sec:exp-setting}
\begin{figure}[h]
    \centering
    \includegraphics[width=0.9\linewidth]{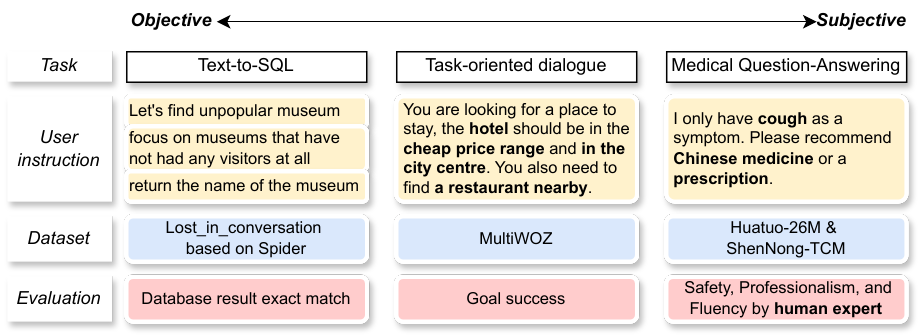}
    \caption{The summary of our experiment tasks.}
    \label{fig:experiments}
\end{figure}

In this study, we focus on iterative meta-prompting by leveraging textual feedback from the environment. 
We conduct experiments on three challenging human–machine interaction tasks that require multiple turns: Text-to-SQL, Task-oriented Dialogue, and Medical Question-answering (Section~\ref{sec:tasks}). 
An overview is shown in \autoref{fig:experiments}. 
Our meta-prompting components are \emph{task-agnostic} (Section~\ref{sec:meta-prompt}). 
They are designed to optimise the prompt of interactive LLM-based systems (Section~\ref{sec:optimisation}).
All prompts are zero-shot in-context instructions\footnote{Following \citet{brown2020languagemodelsfewshotlearners}, this is in-context learning because task descriptions are provided as context, and zero-shot because no demonstrations are included.}.

\subsection{Tasks}
\label{sec:tasks}
\label{sec:tasks}
\paragraph{Text-to-SQL}
\citet{laban2025llmslostmultiturnconversation} proposed 6 tasks to study the performance drop of LLMs from fully-specified user queries to multi-turn interactions.
The multi-turn, sharded instruction (e.g., Shard 1 conveys the high-level intent, and subsequent shards provide incremental clarifications) is partitioned based on the single-turn, fully-specified instruction from the original dataset.
The largest decline occurs in the Text-to-SQL task, which we therefore select to study under different prompt optimisation methods, using instructions and databases from the Spider dataset \citep{yu2018spider}.

In this task, the system agent receives a database schema at the start of the interaction and generates SQL queries from user queries in natural language. 
We evaluate closed-source LLMs (GPT-4o mini, Gemini-2.0-flash) and open-source LLMs (Llama-3.1-8B, Llama-3.1-70B, Llama-4-scout).
The agent is optimised in the multi-sharded environment and evaluated by \emph{functional accuracy}.

\paragraph{Task-oriented Dialogue}
To evaluate on a more realistic scenario, we conduct experiments on MultiWOZ 2.1 \citep{budzianowski-etal-2018-multiwoz,eric-etal-2020-multiwoz}, containing 10k human-to-human conversations on information-seeking, recommendations, and reservations across multiple domains. 
In this work, we focus on the attraction, hotel, restaurant, and train domains, under the ConvLab-3 framework \citep{zhu-etal-2023-convlab}. 
Each user goal of the simulated user is a plain-text description, e.g., \emph{``You are looking for a place to stay, the hotel should be in the cheap price range and in the city centre. You also need to find a restaurant nearby.''}  

The system agent is FnCTOD~\citep{li-etal-2024-large-language-models}, built with GPT-4o mini. 
FnCTOD has two prompts (state tracking via function calls and response generation), both optimised in our setting.
Performance is measured by \emph{success rate} using the rule-based evaluator in ConvLab-3.

\paragraph{Medical Question-Answering}
To evaluate our system in a more human-centred setting and how well prompting can improve the model's performance in a domain that is not common in the pre-training data, we use two medical question-answering datasets: Huatuo-26M \citep{li2023huatuo} and ShenNong-TCM \citep{zhu2023ChatMed}, which are collected from the internet, e.g., encyclopedias, books,
literature, and web corpus, or generated by an LLM based on a traditional Chinese medicine entity graph in Huatuo-26M and ShenNong-TCM, respectively.
 Simulated users act based on descriptions in plain text, related to general medicine or traditional Chinese medicine, e.g., \emph{``\begin{CJK*}{UTF8}{bsmi}我食欲不振，如何改善？\end{CJK*}(I am experiencing a loss of appetite. How to get it back?)''}.

The system agent is built with GPT-4o mini and relies solely on pre-training knowledge (no external knowledge base).
At each epoch, an expert with degrees in general medicine and traditional Chinese medicine provides feedback on 10 interactions. 
For evaluation, three experts compare system pairs on 90 interactions per domain (general medicine and traditional Chinese medicine), based on safety, professionalism, and fluency \citep{yang2024zhongjing}.

\subsection{Meta-prompting components}
\label{sec:meta-prompt}
In interactive optimisation, the feedbacker $\LLM_F$ and rewriter $\LLM_R$ use either closed-source LLMs (GPT-4o mini, Gemini-2.0-flash) or open-source LLMs (Llama-3.1-8B, Llama-3.1-70B, Llama-4-scout). 
More detail is shown in \autoref{tab:model_list}.
Across different tasks, the prompts of $\LLM_F$ and $\LLM_R$ remain fixed, highlighting the task-independent role of these components.

\subsection{Optimisation and Evaluation}
\label{sec:optimisation}
We first collect interactions using the initial prompt and user instructions sampled from the training set. 
Due to context limits and to facilitate human feedback, the feedbacker processes 10 interactions per epoch.  
The rewriter then generates 2 new prompts based on the previous prompt and the feedback. 
New interactions are collected with each candidate prompt, and the one with the highest score on the validation set (based on automatic metrics or human experts, depending on the task) is chosen for the next iteration.  
We use $5$ epochs for Text-to-SQL, $8$ for Task-oriented Dialogue, and $3$ for Medical Question-Answering.
\paragraph{Baselines} 
In our experiments, we compare various prompt optimisation methods.
\emph{Black-Box Prompt Optimization (BPO)} fine-tunes Llama-2-7B-Chat \citep{touvron2023llama} for prompt rewriting based on preference learning \citep{cheng-etal-2024-black}.
\emph{Automatic Prompt Optimisation (APO)} uses the user input, system output, and label to generate feedback \citep{pryzant-etal-2023-automatic}. 
For multi-turn interactions, golden labels are infeasible since multiple solution paths exist; thus, we use a binary success/failure label.
\emph{Gradient-inspired Prompt Optimizer (GPO)} iteratively updates prompts via numerical feedback, e.g., functional accuracy for Text-to-SQL, task success for dialogue \citep{tang2024unleashing}.
\emph{MC-style (TextGrad) }\citep{yuksekgonul2025optimizing} processes entire conversations to generates textual feedback (see Section~\ref{sec:feedback-generation}).

%% file: tex/4_results.tex
\section{Results and Discussion}
\label{sec:results-and-discussion}

\subsection{Robustness and Generalisability}
\label{sec:robustness}
\paragraph{System agents as different LLMs} \autoref{tab:optimise-different-llm} shows the results of optimising system agents built on five LLM backbones for the text-to-SQL task. 
Prompt optimisation methods aim to improve system agents in the multi-sharded setting, i.e., the user only reveals part of the information in one turn.
For comparison, $\text{Oracle}_{\texttt{Full}}$, a single-turn setting where the user query is fully specified at once, is taken as an upper bound.
The performance gap between $\text{Baseline}_{\Sharded}$ and $\text{Oracle}_{\texttt{Full}}$ (average $0.333$ vs. $0.743$) highlights the difficulty LLMs face in handling multi-turn interactive tasks.

The prompts optimised by BPO do not improve performance; in some cases, they even degrade it, especially when applied to systems built on different model families, e.g. GPT-4o mini or Gemini-2.0-flash. 
It is not surprising that BPO is inferior since it is optimised for revising prompts for single-turn tasks and does not generalise well to multi-turn scenarios. Its generalisability is further limited by its dependence on the Llama family as the backend, resulting in suggestions that may not transfer to GPT- or Gemini-based system agents. On the other hand, $\text{RPO}_\text{TD}$ outperforms prior approaches when the system agent is built with Gemini-2.0-flash, Llama-4-scout, and Llama-3.1-70B. 
In contrast, $\text{RPO}_\text{TD+replay}$ achieves the best overall performance, with an average score of $0.477$ ($+54.2\%$ over $\text{Baseline}_{\Sharded}$). 
Llama-3.1-8B benefits the most, since its performance optimised by $\text{RPO}_\text{TD+replay}$ ($0.467$) nearly matches the oracle fully-specified setting ($0.505$). 
The gains across both closed-source and open-source backbones show strong robustness.

However, despite substantial gains over the sharded baseline, a gap to the baseline with the fully-specified user query (average $0.477$ vs. $0.743$) underscores that prompt optimisation can mitigate, but not fully eliminate, the degradation caused by multi-turn interactions.

\begin{table*}[ht]
\small
\centering
\caption{
Functional accuracy with a 95\% confidence interval of Text-to-SQL system agents built on five LLMs optimised with various methods. $\text{Oracle}_{\texttt{Full}}$: An oracle baseline in a single-turn setting with fully-specified user queries. 
The final two columns show the average score (Mean) and the relative improvement ($\Delta\%$) over the $\text{Baseline}_{\Sharded}$.
\textbf{Bold} scores are significantly better than the others ($p < 0.05$).}
\label{tab:optimise-different-llm}
\begin{tabular}{@{}l@{\hspace{2pt}}c@{\hspace{7pt}}c@{\hspace{7pt}}c@{\hspace{7pt}}c@{\hspace{7pt}}ccr@{}}
\toprule
\multirow{2}{*}{Method} & \multicolumn{5}{c}{LLM of the \emph{system} agent} & \multirow{2}{*}{\textbf{Mean}} & \multirow{2}{*}{$\bf \Delta\%$} \\ \cmidrule(lr){2-6}
 & GPT & Gemini & Llama-4 & Llama-8B & Llama-70B &  &  \\ \midrule
$\text{Baseline}_{\Sharded}$ & $0.402_{\pm 0.002}$ & $0.514_{\pm 0.027}$ & $0.206_{\pm 0.013}$ & $0.224_{\pm 0.016}$ & $0.318_{\pm 0.014}$ & $0.333$ & - \\
BPO & $0.318_{\pm 0.061}$ & $0.397_{\pm 0.069}$ & $0.220_{\pm 0.026}$ & $0.185_{\pm 0.028}$ & $0.241_{\pm 0.061}$ & $0.272$ & $-15.7$ \\
APO  & $0.374_{\pm 0.009}$ & $0.523_{\pm 0.008}$ & $0.318_{\pm 0.012}$ & $0.290_{\pm 0.013}$ & $0.336_{\pm 0.004}$ & $0.368$ & $16.9$ \\
GPO & $0.458_{\pm 0.008}$ & $0.523_{\pm 0.016}$ & $0.299_{\pm 0.017}$ & $0.290_{\pm 0.015}$ & $0.308_{\pm 0.011}$ & $0.376$ & $17.5$ \\
MC-style  & $0.459_{\pm 0.018}$ & $0.551_{\pm 0.015}$ & $0.250_{\pm 0.013}$ & $0.346_{\pm 0.021}$ & $0.332_{\pm 0.007}$ & $0.388$ & $20.4$ \\ \midrule
$\text{RPO}_\text{TD}$ (ours) & $0.439_{\pm 0.011}$ & $0.561_{\pm 0.013}$ & $0.336_{\pm 0.014}$ & $0.318_{\pm 0.009}$ & $0.383_{\pm 0.011}$ & $0.408$ & $28.9$ \\
$\text{RPO}_\text{TD+replay}$ (ours)& $\textbf{0.528}_{\pm 0.011}$ & $\textbf{0.607}_{\pm 0.018}$ & $\textbf{0.383}_{\pm 0.013}$ & $\textbf{0.467}_{\pm 0.022}$ & $\textbf{0.402}_{\pm 0.012}$ & $\textbf{0.477}$ & $\bf 54.2$ \\ \midrule
$\text{Oracle}_{\texttt{Full}}$ & $0.893_{\pm 0.007}$ & $0.841_{\pm 0.007}$ & $0.729_{\pm 0.017}$ & $0.505_{\pm 0.017}$ & $0.748_{\pm 0.014}$ & $0.743$ & $140.2$ \\ \bottomrule
\end{tabular}
\end{table*}

\paragraph{Prompt optimisation with different LLMs}
\label{sec:ablation-RPO}
\autoref{tab:multiwoz-result} reports the success rates of FnCTOD ~\citep{li-etal-2024-large-language-models} when optimised by different prompt optimisation methods across five LLM backbones. 
The baseline system achieves a success rate of $0.420$, while all optimisation methods substantially improve performance. 
Among prior approaches, MC-style feedback yields the strongest results with a mean success rate of $0.565$ ($+34.4\%$ over baseline), slightly outperforming APO and GPO. 
Our proposed methods consistently surpass these baselines. 
In particular, $\text{RPO}_{\text{TD}}$ achieves a mean score of $0.575$ ($+37.0\%$), demonstrating the advantage of trajectory-driven optimisation. 
When combined with the rewriter with experience replay, $\text{RPO}_{\text{TD+replay}}$ delivers the best performance across all LLMs, reaching an average success rate of $0.619$, corresponding to a relative improvement of 47.3\%. 
The gains are consistent across all five LLMs, confirming that RPO is robust and generalisable, independent of the underlying model of the meta-prompting agents.
\begin{table*}[ht]
\small
\centering
\caption{The success rate with a 95\% confidence interval of the task-oriented dialogue system, FnCTOD~\citep{li-etal-2024-large-language-models}, improved by various prompt optimisation methods leveraging 5 different LLMs. 
The initial success rate of FnCTOD is $0.420$. 
\textbf{Bold} scores are significantly better than the others ($p < 0.05$).}
\label{tab:multiwoz-result}
\begin{tabular}{@{}l@{\hspace{2pt}}c@{\hspace{7pt}}c@{\hspace{7pt}}c@{\hspace{7pt}}c@{\hspace{7pt}}ccr@{}}
\toprule
\multirow{2}{*}{Method} & \multicolumn{5}{c}{LLM of the \emph{meta-prompting} agent} & \multirow{2}{*}{\textbf{Mean}} & \multirow{2}{*}{$\bf \Delta\%$} \\ \cmidrule(lr){2-6}
 & GPT & Gemini & Llama-4 & Llama-8B & Llama-70B &  &  \\ \midrule
APO & $0.540_{\pm 0.030}$ & $0.560_{\pm 0.018}$ & $0.540_{\pm 0.018}$ & $0.560_{\pm 0.018}$ & $0.560_{\pm 0.018}$ & $0.552$ & $31.4$ \\
GPO  & $0.579_{\pm 0.021}$ & $0.541_{\pm 0.008}$ & $0.571_{\pm 0.046}$ & $0.554_{\pm 0.017}$ & $0.526_{\pm 0.053}$ & $0.554$ & $32.0$ \\
MC-style  & $0.567_{\pm 0.042}$ & $0.549_{\pm 0.010}$ & $0.575_{\pm 0.038}$ & $0.560_{\pm 0.029}$ & $0.572_{\pm 0.044}$ & $0.565$ & $34.4$ \\ \midrule
$\text{RPO}_\text{TD}$ (ours) & $0.578_{\pm 0.037}$ & $0.562_{\pm 0.036}$ & $0.586_{\pm 0.021}$ & $0.594_{\pm 0.013}$ & $0.556_{\pm 0.042}$ & $0.575$ & $37.0$ \\
$\text{RPO}_\text{TD+replay}$ (ours) & $\textbf{0.625}_{\pm 0.038}$ & $ \textbf{0.622}_{\pm 0.018}$ & $\textbf{0.618}_{\pm 0.018}$ & $\textbf{0.622}_{\pm 0.007}$ & $0.606_{\pm 0.020}$ & $\bf 0.619$ & $\bf 47.3$ \\ \bottomrule
\end{tabular}
\end{table*}

\begin{figure*}[h]
    \centering
    \begin{subfigure}[b]{0.475\textwidth}
        \centering
        \includegraphics[width=\linewidth]{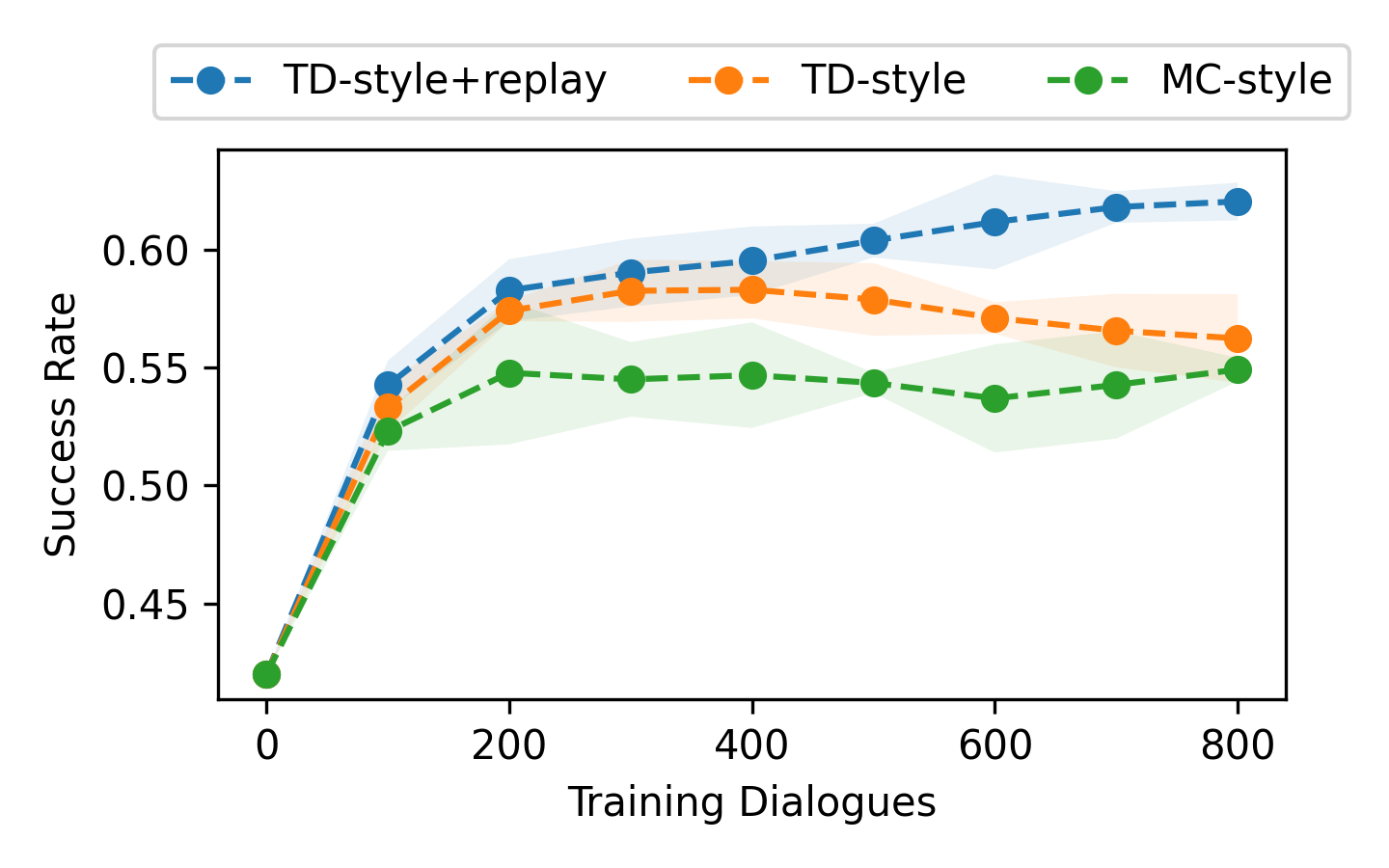}
    \caption{Different textual feedback generation methods.}
    \label{fig:gemini-dialog-train}
    \end{subfigure}%
    \begin{subfigure}[b]{0.475\textwidth}
        \centering
        \includegraphics[width=\linewidth]{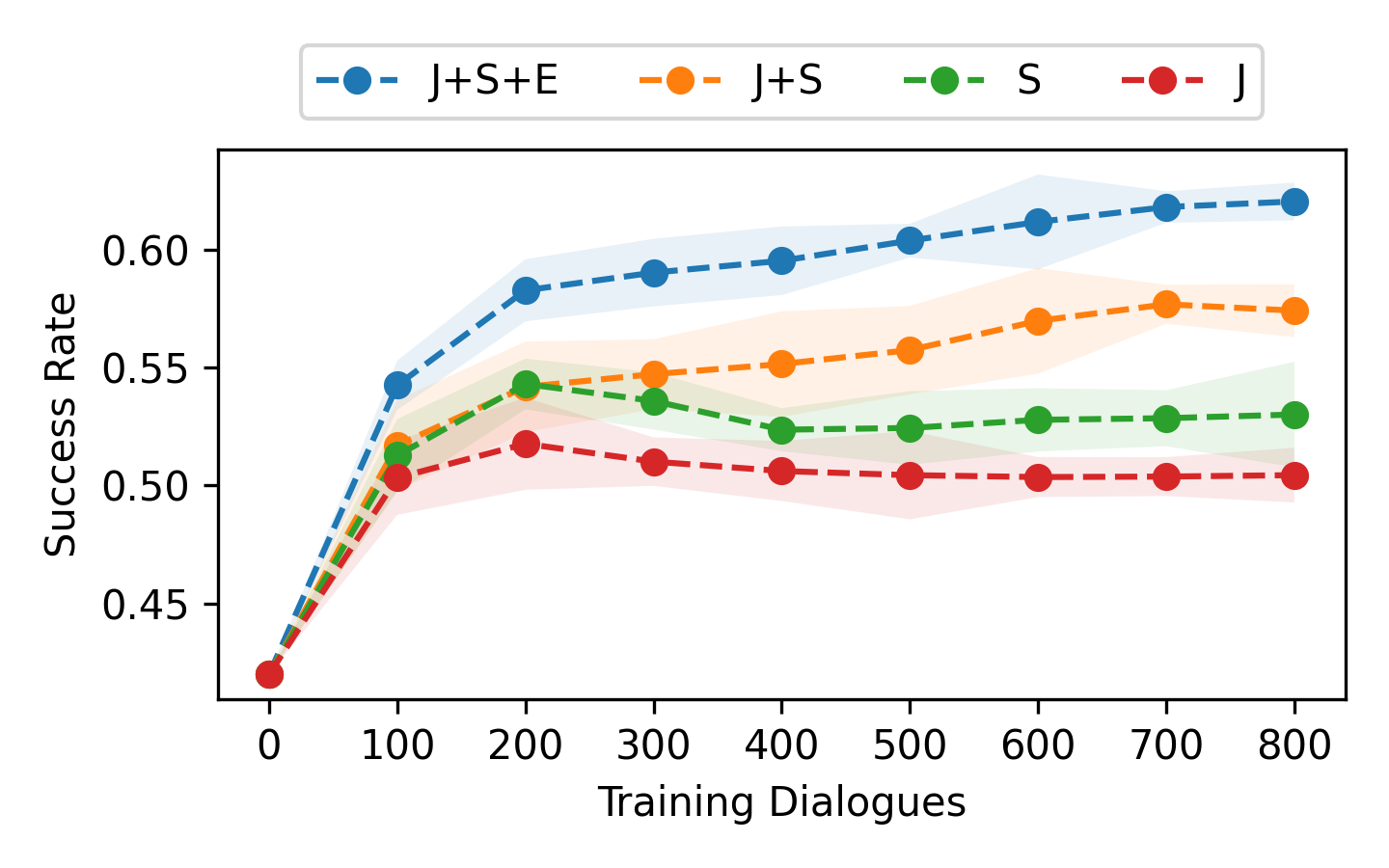}
    \caption{Ablation on the feedbacker's prediction fields.}
    \label{fig:ablation-different-prediction}
    \end{subfigure}%
    \hfill
    \begin{subfigure}[b]{0.475\textwidth}
        \centering
        \includegraphics[width=\linewidth]{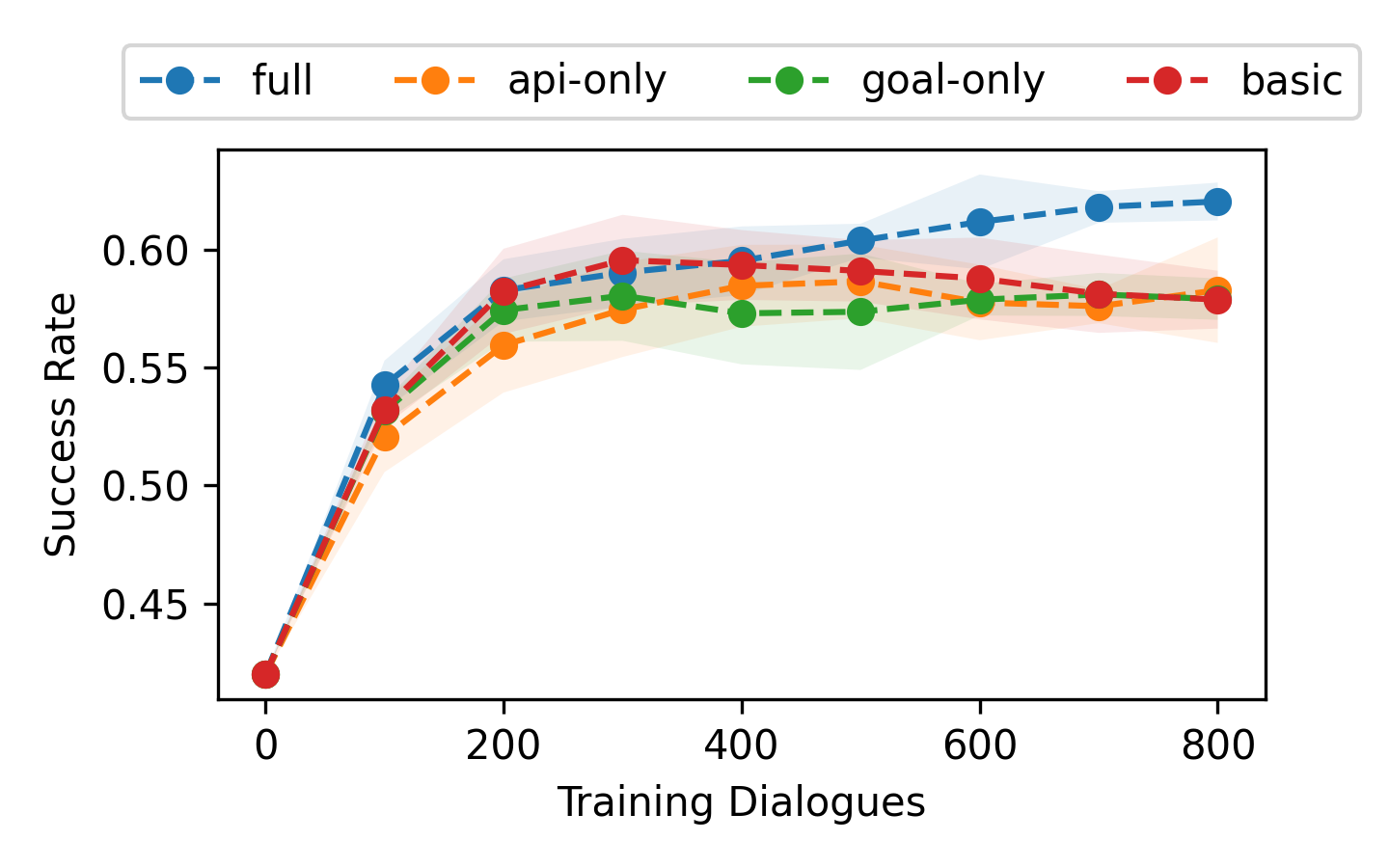}
    \caption{Different info for RPO. }
    \label{fig:ablation}
    \end{subfigure}%
    \begin{subfigure}[b]{0.475\textwidth}
        \centering
        \includegraphics[width=\linewidth]{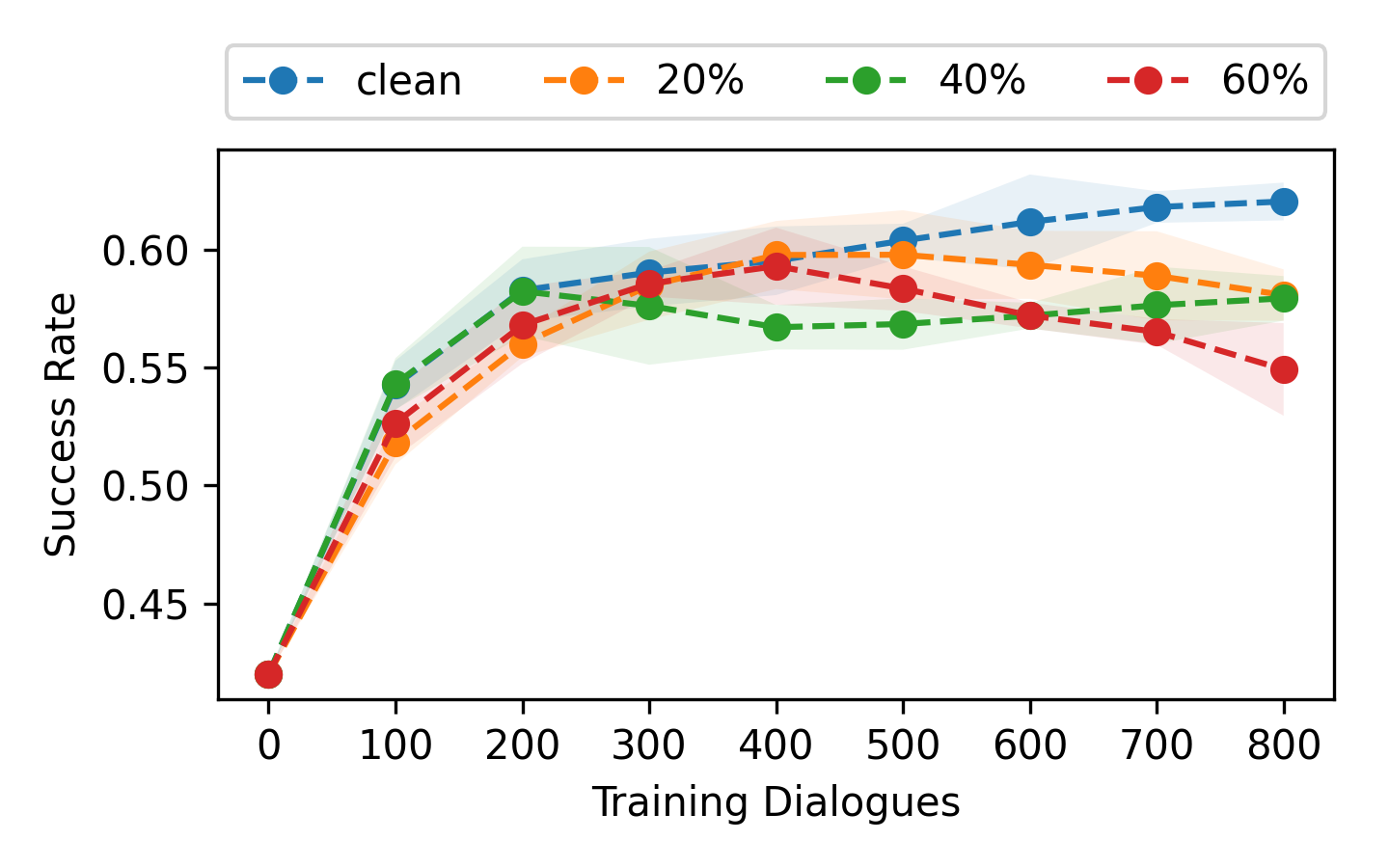}
    \caption{Noisy external verifiable evaluation signal.}
    \label{fig:noisy}
    \end{subfigure}
    \caption{
         Each setting uses 4 training seeds and is evaluated on 100 dialogues; the line represents the mean success rate, and the shaded area indicates the standard error.
    }
\label{fig:ablations}
\end{figure*}

\subsection{Ablation of feedback design in RPO}
\label{sec:ablation-different-feedback}

To examine whether the gains come from richer feedback design rather than generic iterative prompt refinement, we conduct four ablations:  feedback style (MC vs.\ TD; \autoref{fig:gemini-dialog-train}), the prediction fields generated by the feedbacker (\autoref{fig:ablation-different-prediction}), input signals provided to the feedbacker (\autoref{fig:ablation}), and robustness to noisy external feedback (\autoref{fig:noisy}).

\paragraph{Feedback style and training curves} 
The training curves of FnCTOD optimised by MC-style, TD-style, and TD-style+replay with Gemini-2.0-flash are shown in \autoref{fig:gemini-dialog-train} (results for other LLMs are shown in \autoref{fig:coverage}).
MC-style exhibits higher variance during early training, whereas TD-style is more stable and converges faster.
With further training, their final performances become comparable.
Incorporating experience replay into the rewriter further stabilises training and achieves the best overall performance. Qualitative comparisons of MC- and TD-style feedback, along with the resulting optimised prompts, are provided in \autoref{app:feedbacks}.

\paragraph{Prediction targets and performance gains} 
We further ablate which \emph{prediction fields} drive the gains, e.g. whether the short-term reward (next-turn user emotion) and long-term value (future task success) help.
As shown in \autoref{fig:ablation-different-prediction}, no single target accounts for the observed gains: using only \emph{J} or only \emph{S} underperforms their combination (\emph{J+S}), while the full setting (\emph{J+S+E}) achieves the best performance. These results indicate that richer and more structured prediction targets are critical for effective feedback, beyond generic iterative prompt rewriting.

\paragraph{Effect of input signals on feedback quality} 
We examine the impact of input signals provided to the feedbacker. 
The \emph{basic} setting uses only the dialogue text. 
The \emph{goal-only} setting additionally provides the user goal, while the \emph{api-only} setting includes the system API call. The \emph{full} setting (TD-style with replay) incorporates both signals. 
As shown in \autoref{fig:ablation}, both the user goal and the API call are essential for optimal performance.
Although these signals can often be inferred implicitly, providing them explicitly leads to substantial improvements. 
This suggests that direct access to task-relevant state information enables the feedbacker to generate more specific and actionable feedback.
Example prompts before and after optimisation with $\text{RPO}_{\text{TD+replay}}$ are shown in \autoref{fig:dialog-before-training} and \autoref{fig:dialog-after-training}.

\paragraph{Robustness to noisy feedback} 
To more substantiate the impact of the \emph{quality of textual feedback}, we flipped $20\%$, $40\%$, and $60\%$ of the external evaluation signal to assess robustness to noisy external evaluation signals. 
As shown in \autoref{fig:noisy}, the performance is robust when the noise level is up to $40\%$. The optimisation ability of RPO declines when the noise reaches $60\%$, which reverses the reward signal on average and naturally harms any optimisation. On the other hand, if the noise level is less than $50\%$, its performance is close to the clean setting. In other words, the turn-level feedback, which is not conditioned on the external feedback, provides a learning signal that, to some extent, mitigates the noise in the external evaluation signal.

\subsection{Prompting limitations on underrepresented topics in LLMs}
\label{sec:prompting-limitation}
\begin{figure*}[h]
    \centering
    \begin{subfigure}[b]{0.5\textwidth}
        \centering
        \includegraphics[width=0.9\textwidth]{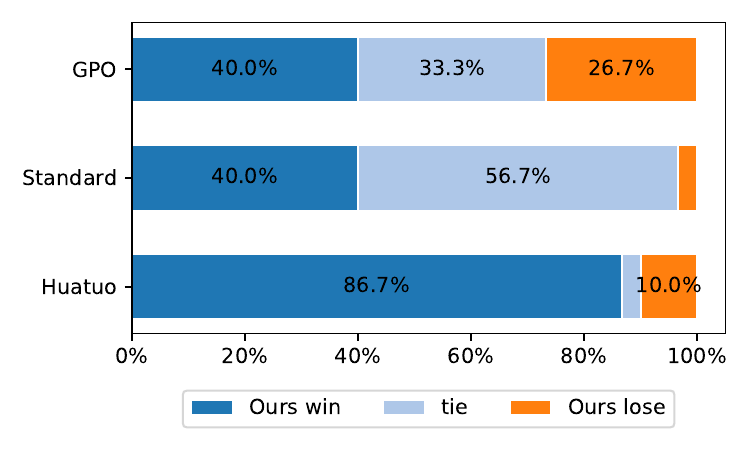}
        \caption{Result on general medicine.}
        \label{fig:medqa-general}
    \end{subfigure}%
    \begin{subfigure}[b]{0.5\textwidth}
        \centering
        \includegraphics[width=0.9\textwidth]{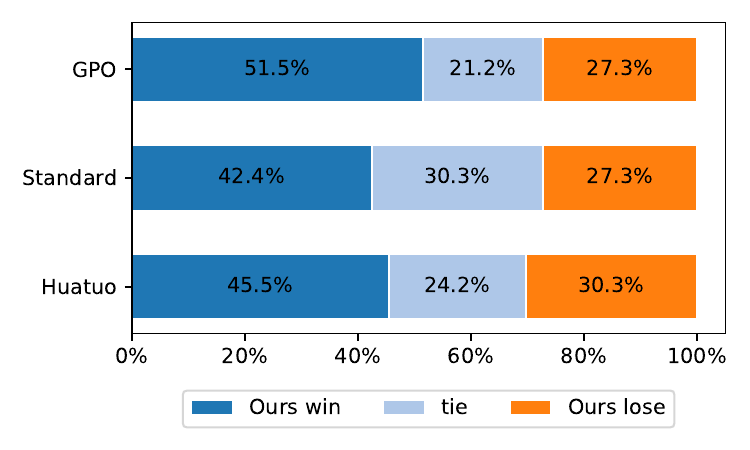}
        \caption{Results on traditional Chinese medicine.}
        \label{fig:med-qa-tcm}
    \end{subfigure}
    \caption{
        Overall preference between our method and a standard system (Standard), GPO, and Huatuo{GPT}-{II} (Huatuo) on the medical question-answering task.
        The overall recommendation by human experts is based on safety, professionalism, and fluency.
    }
    \label{fig:med-qa-result}
\end{figure*}

We compare our method against three systems: a standard system, built with GPT-4o mini with the initial prompt, a standard system updated via GPO, and HuatuoGPT-II~\citep{chen2024huatuogptii}, a large language model which is fully fine-tuned on medical data and demonstrates the state-of-the-art performance on Chinese medicine benchmarks. In other words, except Huatuo{GPT}-{II}, a fully fine-tuned 7B model, all systems are built with GPT-4o mini by prompting.

In general medicine, our method consistently outperforms the fully fine-tuned Huatuo{GPT}-{II} with an 86.7\% win rate and is preferred over other prompting-based baselines (see \autoref{fig:medqa-general}). 
On the other hand, traditional Chinese medicine is more challenging. For example, our system's preference rate drops by 41\% compared to Huatuo when transitioning from general medicine to traditional Chinese medicine (see \autoref{fig:med-qa-tcm}).
However, despite this drop in preference, our proposed method is still favoured in general.

This observation is aligned with the findings by \citet{petrov2024when}. Our method performs better in general medicine because the skills present in the pre-training data of LLMs can be elicited by prompting. 
However, tasks that are unseen or underrepresented in pre-training data are hard to learn through prompting. 
How to properly leverage external knowledge to improve the performance on unseen or under-represented tasks is an important future work.

%% file: tex/5_conclusion.tex
\section{Conclusions}
\label{sec:conclusions}
We proposed a robust framework for interactive prompt optimisation that can effectively optimise system agents built on diverse LLM backbones and system architectures, ranging from standard input–output agents in text-to-SQL and medical QA to multi-stage agents in task-oriented dialogue that access external knowledge sources.
In addition, it is flexible to the choice of LLM used for generating feedback and rewriting, as it works effectively with both closed-source LLMs (GPT-4o mini and Gemini-2.0-flash) and open-source LLMs (Llama variants).
Turn-level feedback enriched with user status and API details, together with experience replay in rewriting, proved highly effective for stabilising and enhancing optimisation in multi-turn tasks.

By using the optimised prompt, the system can minimise the need for extensive self-feedback loops, reducing computational overhead and API call frequency during inference.
Although the performance optimised by our method still falls short of fully specified settings and unseen tasks remain difficult to optimise purely by prompting, our reinforcement learning-inspired method offers a stable, practical, and efficient approach for automatic prompt optimisation to reduce the challenges of unspecified multi-turn interactions, which could be valuable for future LLM research.

%% file: tex/appendix/F_limitations.tex
\section{Limitations}
\label{appx:limitations}
\paragraph{Limitation in the learnability} Although our method is able to adapt to various tasks, RPO's performance can be limited for skills that are underrepresented in the pretraining data, e.g. traditional Chinese medicine, as mentioned in Section~\ref{sec:prompting-limitation}. 
\paragraph{Limitation in the reproducibility} Although we release our code, datasets, and hyperparameters, exact reproduction of the results may not be guaranteed.
LLMs accessed via API are not batch-invariant, meaning their behaviour can vary with different batch sizes, and this inherent nondeterminism introduces unavoidable variability~\citep{he2025nondeterminism}.
Further details are provided in \autoref{sec:training}. 

%% file: tex/appendix/A_model_list.tex
\section{Compute resource and Model list}
\label{sec:llm}
\begin{table}[h]
\centering
\caption{Specific model versions used in our experiments. }
\label{tab:model_list}
\begin{tabular}{@{}llll@{}}
\toprule
Short Form & Name & Version & Access Provider \\ \midrule
GPT & GPT-4o mini & gpt-4o-mini-2024-07-18 & OpenAI \\
Gemini & Gemini-2.0-flash & gemini-2.0-flash-001 & VertexAI \\
Llama-4 & Llama-4-scout-17B-16E & llama-4-scout-17b-16e-instruct-maas & VertexAI \\
Llama-8B & Llama-3.1-8B & N/A & VertexAI \\
Llama-70B & Llama-3.1-70B & N/A & VertexAI \\ \bottomrule
\end{tabular}
\end{table}
\paragraph{Cost of RPO}
The execution time of RPO varies depending on the task. 
For instance, generating feedback per trajectory takes $5.91 \pm 0.09$ seconds for Text-to-SQL and $6.94 \pm 0.07$ seconds for task-oriented dialogue.
Per epoch, RPO consumes $1522.58 \pm 137.2$ tokens, which is comparable to APO ($1423.9 \pm 45.5$) but higher than GPO ($153.0 \pm 29.4$). 
This additional cost reflects a trade-off for obtaining a more informative optimisation signal.

\paragraph{Model List}
The LLMs used in our experiments are listed in \autoref{tab:model_list}. 
All experiments are conducted on Google Cloud Platform.
As the models are accessed via API calls (through OpenAI or Vertex AI; see \autoref{tab:model_list}), the computational requirements are modest.
Specifically, we use a standard virtual machine with 4 vCPUs and 15 GB of memory (machine type \texttt{n1-standard-4}).


%% file: tex/appendix/D_RPO_algorithm.tex
\section{RPO algorithm}
\label{appx:rpo-alg}
The pseudo-code of RPO is given in Algorithm~\ref{alg:rpo}.
\begin{algorithm}

\caption{Reinforced Prompt Optimisation (RPO)}
\label{alg:rpo}
\begin{algorithmic}[1]

\STATE \textbf{Input:}
initial prompt $\prompt^{1}$;
training data $\mathcal{D}_{\text{train}}$;
validation data $\mathcal{D}_{\text{val}}$;
\STATE \textbf{Input:}
system $\LLM_{\text{sys}}$;
user $\LLM_{\text{usr}}$;
\STATE \textbf{Input:}
external evaluator $\texttt{Eval}()$;
simulation environment $\texttt{Interact}()$;
\STATE \textbf{Input:}
feedbacker $\LLM_F$;
rewriter $\LLM_R$;
\STATE \textbf{Input:}
number of epochs $N$;
trajectories $n_{\text{train}}, n_{\text{val}}$;
rewritten prompt candidates $n_p$.

\FOR{$i = 1$ to $N$} 

\STATE
\STATE \textbf{Phase 1: Trajectory Collection}
\FOR{$j = 1$ to $n_{\text{train}}$}
\STATE $t \gets \texttt{Interact}(\LLM_{\text{sys}}(\cdot|\prompt^{i}), \LLM_{\text{usr}}(\cdot|\text{goal}\sim\mathcal{D}_{\text{train}}))$
\STATE $\mathcal{T} \gets \mathcal{T} \cup \{t\}$
\ENDFOR

\STATE
\STATE \textbf{Phase 2: TD-style Feedback Generation}
\FORALL{$t \in \mathcal{T}$}
    \FOR{$t_j$ in trajectory $t$}
        \STATE $\feedback_{\TD,j}^i \gets \LLM_F(t_1, \feedback_{\TD,1}^i, \dots, t_j)$
        \COMMENT{turn-level feedback, Eqn.~\ref{equ:TD_feedback}}
        \label{line:turn-level}
    \ENDFOR
    \STATE $\text{s} \gets \texttt{Eval}(t)$
    \STATE $\feedback_{\TD(t)}^i \gets \LLM_F(\feedback_{\TD,1}^i, \feedback_{\TD,2}^i, \ldots, \text{s})$
    \COMMENT{dialogue-level feedback}
    \label{line:dialog-level}
\ENDFOR
\STATE $\feedback_{\TD}^i \gets \LLM_F(\feedback_{\TD(\mathcal{T})}^i)$
\COMMENT{final feedback}
\label{line:final}

\STATE
\STATE \textbf{Phase 3: Prompt Rewriting and Evaluation}
\FOR{$k = 1$ to $n_p$} 
\STATE Generate new prompts: $\prompt^{i+1,k} \gets \LLM_R(\prompt^{i}, \feedback_{\TD}^i, \cdots, \prompt^{1}, \feedback_{\TD}^1)$
\COMMENT{Eqn.~\ref{equ:replay-rewriter}}

\STATE
\FOR{$j = 1$ to $n_{\text{val}}$}
\STATE $t \gets \texttt{Interact}(\LLM_{\text{sys}}(\cdot|\prompt^{i+1,k}), \LLM_{\text{usr}}(\cdot|\text{goal}\sim\mathcal{D}_{\text{val}}))$
\STATE $\mathcal{T}^{k}_{\text{val}} \gets \mathcal{T}^{k}_{\text{val}} \cup \{t\}$
\ENDFOR
\STATE $\text{success}^k \gets \texttt{Eval}(\mathcal{T}^{k}_{\text{val}})$
\COMMENT{evaluate candidate prompt}
\ENDFOR

\STATE
\STATE \textbf{Phase 4: Policy Update}
\STATE $k^* \gets \arg\max_k \text{success}^k$
\STATE $\prompt^{i+1} \gets \prompt^{i+1,k^*}$

\ENDFOR

\STATE \textbf{Return} $\prompt^{N+1}$

\end{algorithmic}

\end{algorithm}

%% file: tex/appendix/B_converge.tex
\section{Converage analysis}
\label{sec:training}
\begin{figure*}[h]
    \centering
    \begin{subfigure}[b]{0.48\textwidth}
        \centering
        \includegraphics[width=0.9\textwidth]{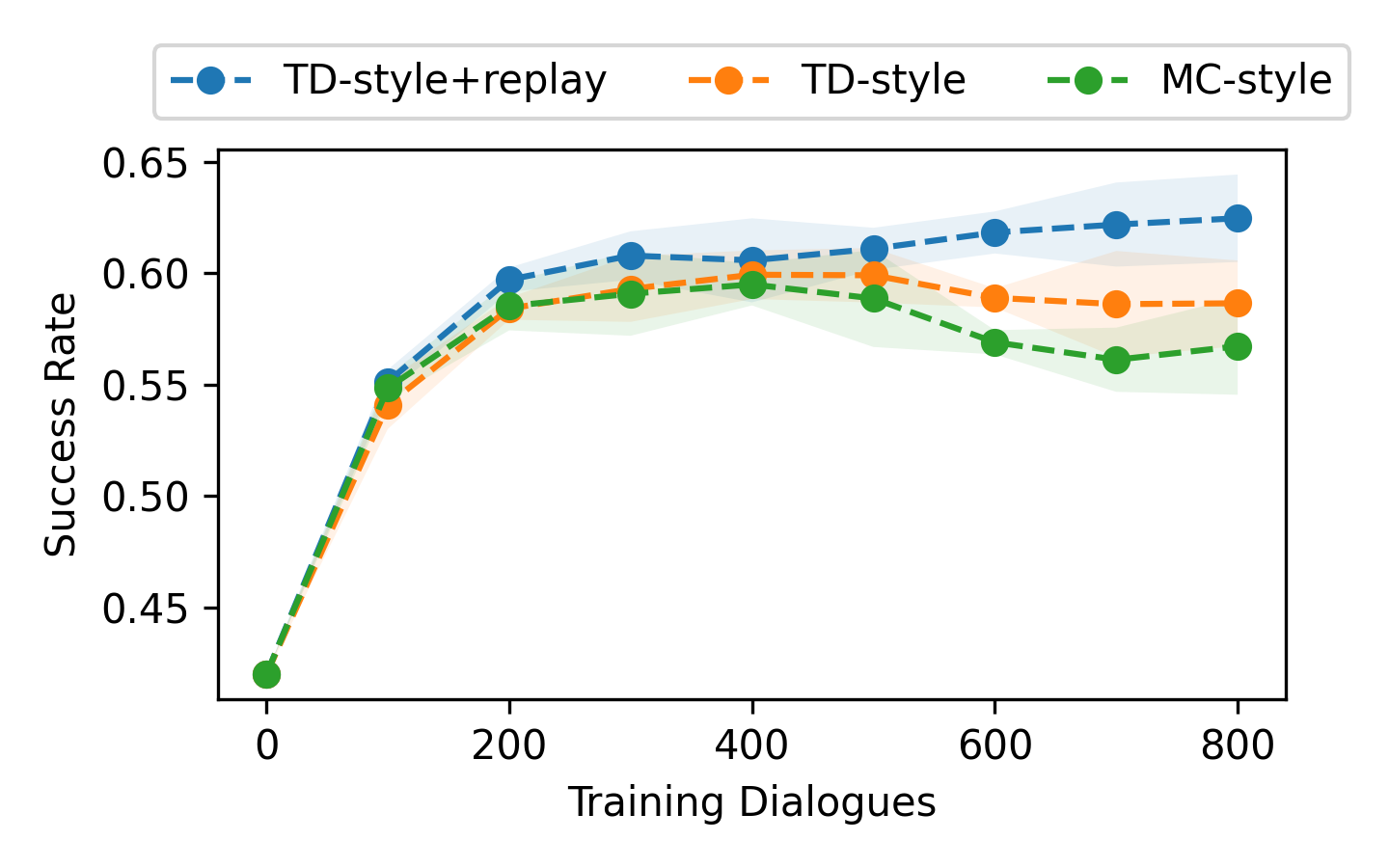}
        \caption{GPT-4o mini}
        \label{fig:gpt}
    \end{subfigure}%
    \begin{subfigure}[b]{0.48\textwidth}
        \centering
        \includegraphics[width=0.9\textwidth]{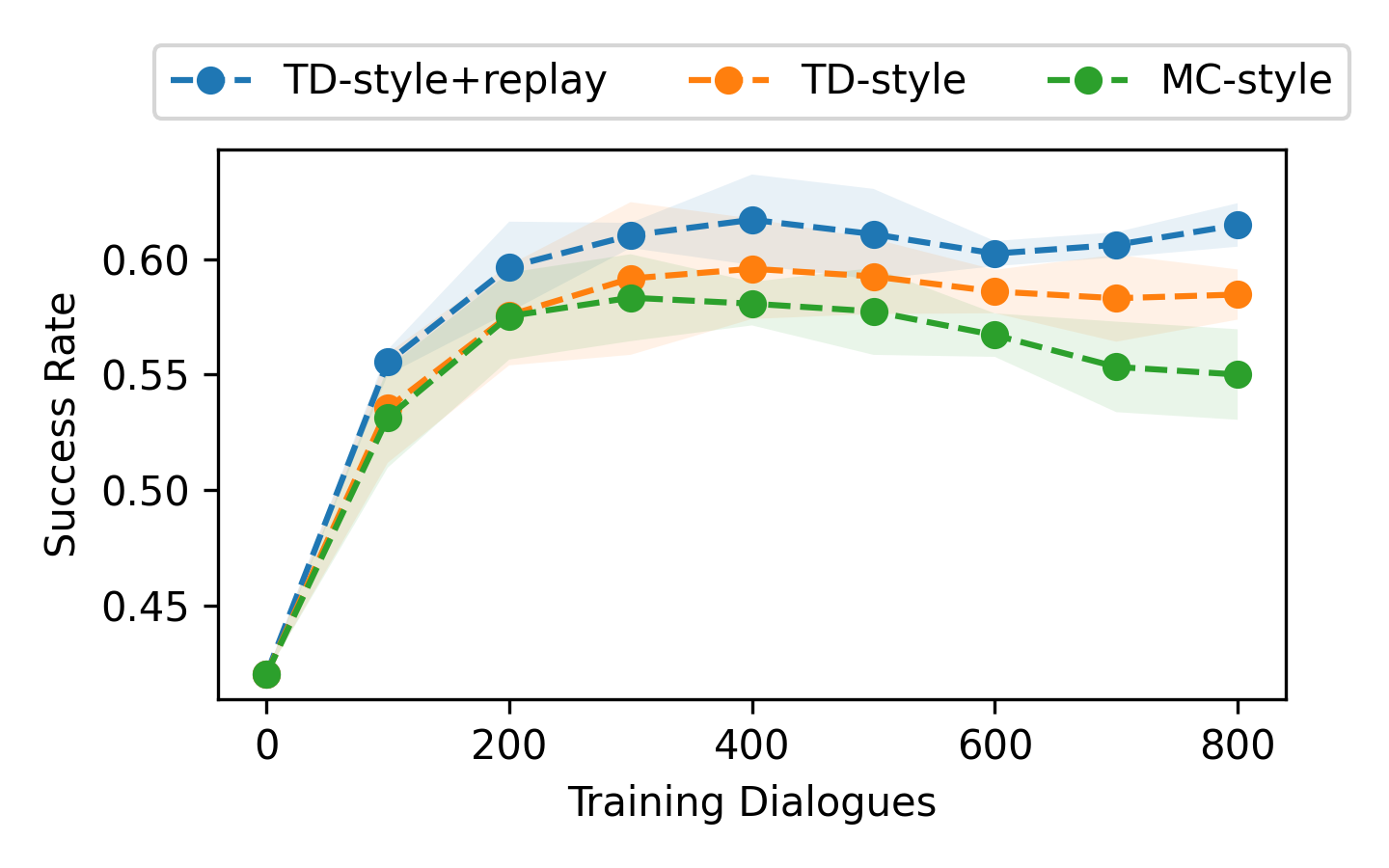}
        \caption{Llama-4-scout}
        \label{fig:llama-4}
    \end{subfigure}
    \begin{subfigure}[b]{0.48\textwidth}
        \centering
        \includegraphics[width=0.9\textwidth]{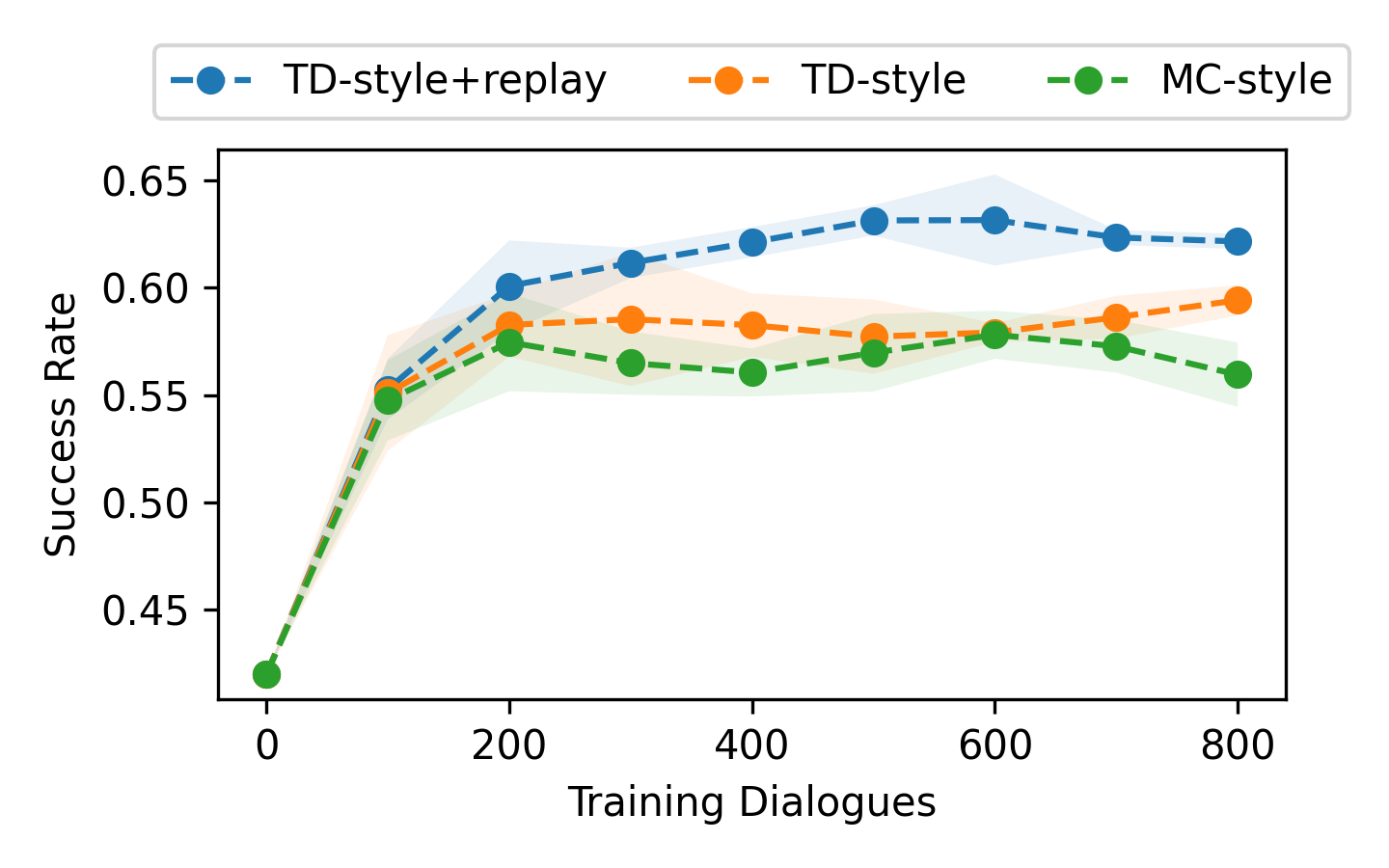}
        \caption{Llama-3.1-8B}
        \label{fig:llama-3b}
    \end{subfigure}%
    \begin{subfigure}[b]{0.48\textwidth}
        \centering
        \includegraphics[width=0.9\textwidth]{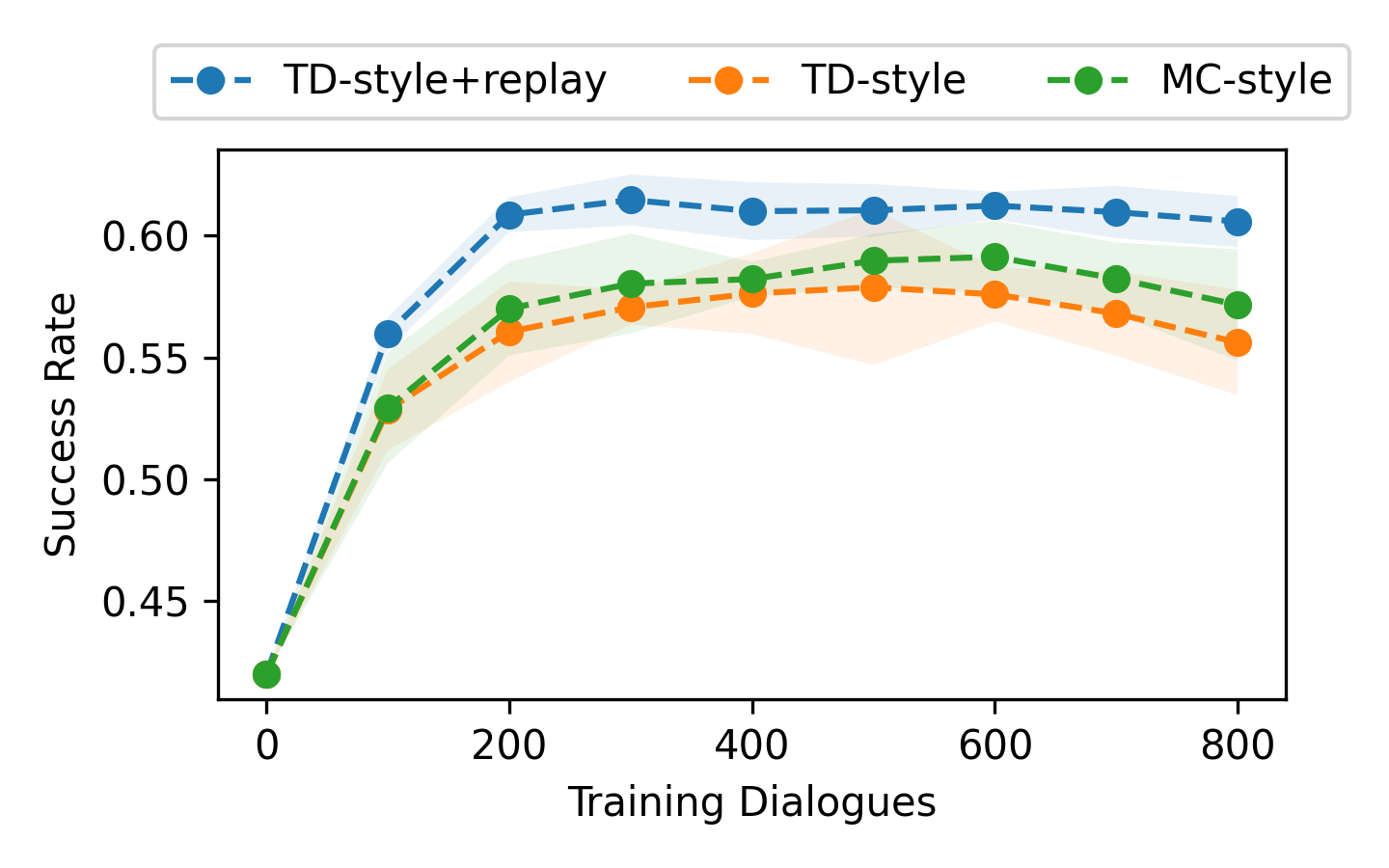}
        \caption{Llama-3.1-70B}
        \label{fig:llama-70b}
    \end{subfigure}
    \caption{
        The training curve of different optimisation methods. Each setting is trained over 4 seeds, evaluated on 100 dialogues. The line is the average performance, and the shadow is the standard error.
    }
    \label{fig:coverage}
\end{figure*}

The training curve of prompt optimisation based on different settings (e.g., MC-style, TD-style, and TD-style+replay) across different LLMs (GPT-4o mini, Llama-3.1-8B, Llama-3.1-70B, and Llama-4-scout) is shown in \autoref{fig:coverage} (The result of Gemini-2.0-flash is shown in \autoref{fig:gemini-dialog-train} previously).

The training curves become stable after epoch 3 (trained with 300 dialogues), and the TD-style+replay setting improves the stability.  
However, since existing LLMs are not batch-invariant, which means their behaviour will be impacted by different batch sizes, there is unavoidable variance caused by their nondeterministic behaviour~\citep{he2025nondeterminism}. 

%% file: tex/appendix/E_qualitative.tex
\section{Qualitative Analysis of TD-style feedback}
\label{app:feedbacks}
\begin{figure*}[h]
    \centering
    \includegraphics[width=0.8\linewidth]{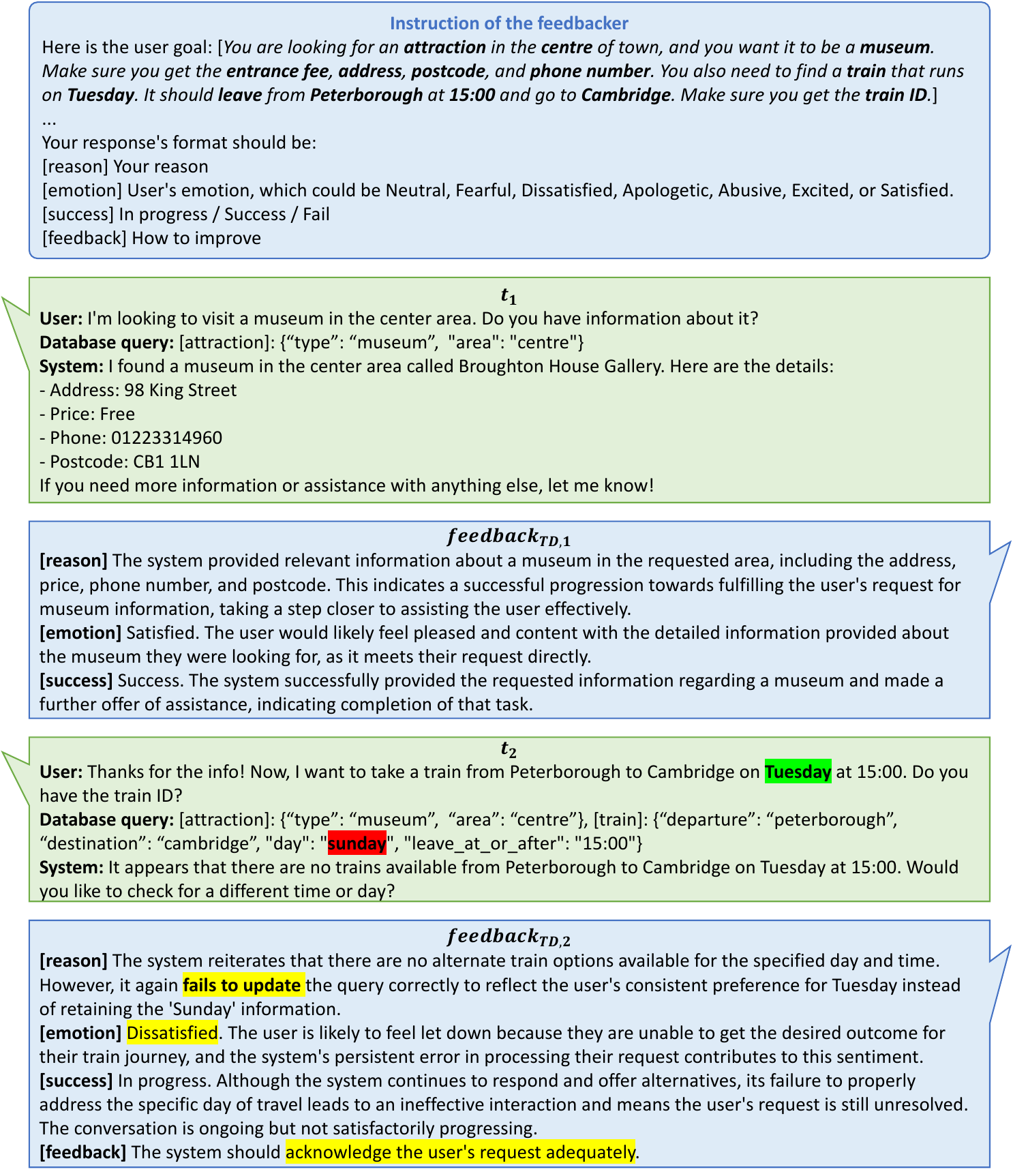}
    \caption{The turn-level feedback generated by the TD-style feedbacker. The first blue block is the instruction of the feedbacker (see full prompt in \autoref{fig:td-feedbacker}). The mistake made by the system is highlighted in red, and the suggestion by the feedbacker is highlighted in yellow, manually.}
    \label{fig:feedback-example}
\end{figure*}

One example of the turn-level feedback generation by $\text{RPO}_{\TD+\text{replay}}$ is shown in \autoref{fig:feedback-example}. 
After receiving the first turn $t_1$, $\feedback_{TD,1}$ is generated, including a positive user emotion estimation and a prediction that the conversation will be successful since there is no mistake at $t_1$. 
However, the system makes a mistake in turn $t_2$, where the API call includes wrong information, i.e. the user mentions ``Tuesday'', but the system puts ``sunday'' in the API call (highlighted in red), resulting in a not found result. $\feedback_{\TD,2}$ points it out, includes negative user emotion prediction, and suggests the system should acknowledge the user's request properly (highlighted in yellow).
This result demonstrates that our proposed feedbacker can estimate the user satisfaction and user success, and provide suggestions properly with reasoning.

\begin{figure*}[h]
    \centering
    \includegraphics[width=0.8\linewidth]{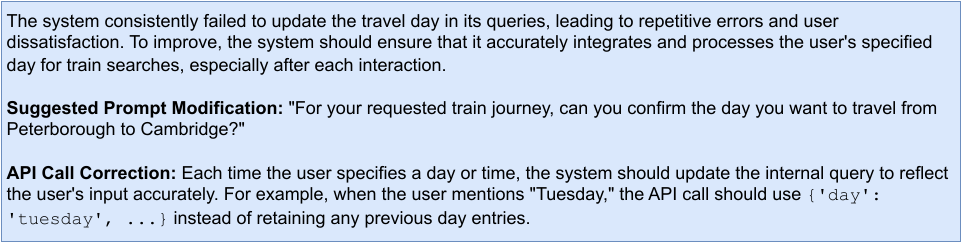}
    \caption{The turn-level feedback in \autoref{fig:feedback-example} will be summarised by the TD-style feedbacker into a dialogue-level feedback.}
    \label{fig:feedback-example-dialogue}
\end{figure*}

The turn-level feedbacks will be summarised by the TD-style feedbacker into a dialogue-level feedback (as shown in \autoref{fig:feedback-example-dialogue} and referred to Line~\ref{line:dialog-level} in Algorithm~\ref{alg:rpo}), and then all dialogue-level feedbacks from the training set will be summarised as the final feedback for this epoch (as shown in \autoref{fig:feedback-example-final} and referred to Line~\ref{line:final} in Algorithm~\ref{alg:rpo}).

\subsection{Comparison with the MC-style feedback} 
\begin{figure*}[h]
    \centering
    \includegraphics[width=0.8\linewidth]{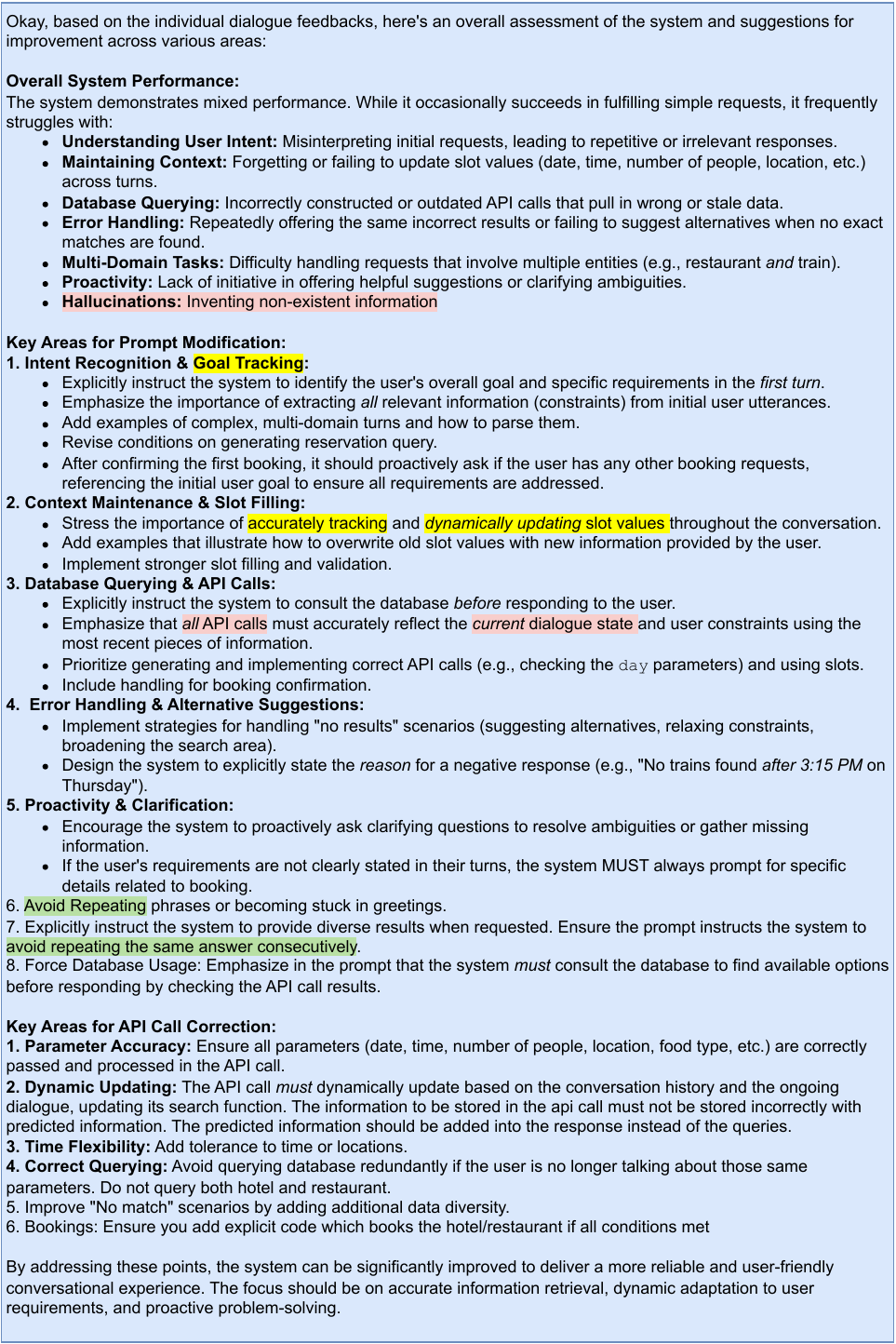}
    \caption{All the dialogue-level feedback, e.g. \autoref{fig:feedback-example-dialogue}, will be further summarised by the TD-style feedbacker into the final feedback $\feedback_\TD$ of this epoch. The suggestions related to goal tracking, API calls, and prevention of looping are manually highlighted in yellow, red, and green, respectively.}
    \label{fig:feedback-example-final}
\end{figure*}

\begin{figure*}[h]
    \centering
    \includegraphics[width=\linewidth]{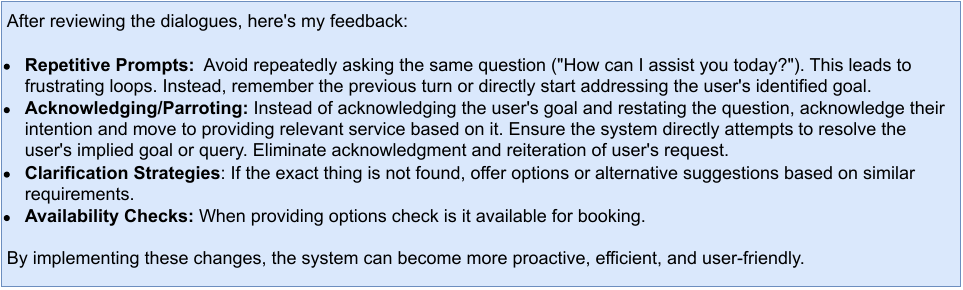}
    \caption{The feedback generated by the MC-style feedbacker.}
    \label{fig:feedback-example-mc-final}
\end{figure*}

The feedback produced by both the TD-style feedbacker (\autoref{fig:feedback-example-final}) and the MC-style feedbacker (\autoref{fig:feedback-example-mc-final}) can offer useful guidance, such as preventing conversational loops or handling cases where information cannot be found. 
However, the TD-style feedbacker captures more fine-grained signals and provides more detailed suggestions, which in turn leads to a faster convergence, as mentioned in Section~\ref{sec:ablation-different-feedback}.

\subsection{The system prompt before and after optimisation}

\begin{figure*}[h]
    \centering
    \includegraphics[width=\linewidth]{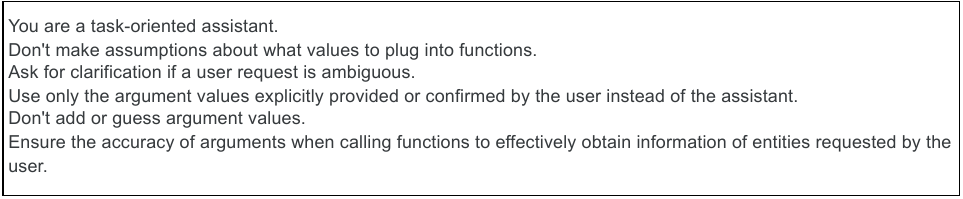}
    \caption{The system prompt of FnCTOD before prompt optimisation.}
    \label{fig:dialog-before-training}
\end{figure*}

\begin{figure*}[h]
    \centering
    \includegraphics[width=\linewidth]{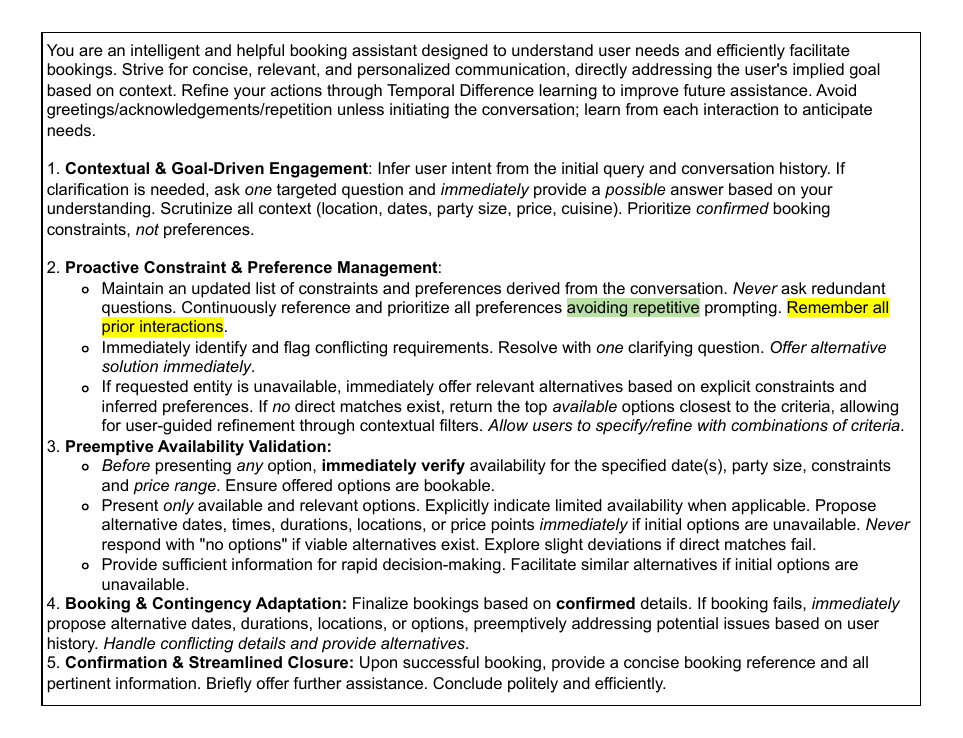}
    \caption{The system prompt of FnCTOD after it is optimised by MC-style for 8 epochs. 
        MC-style is built with Gemini-2.0-Flash. 
        The format is generated by the rewriter in markdown format. 
        For illustration, the instructions of goal tracking (yellow) and looping prevention (green) are manually highlighted.}
    \label{fig:dialog-after-training-mc}
\end{figure*}

\begin{figure*}[h]
    \centering
    \includegraphics[width=\linewidth]{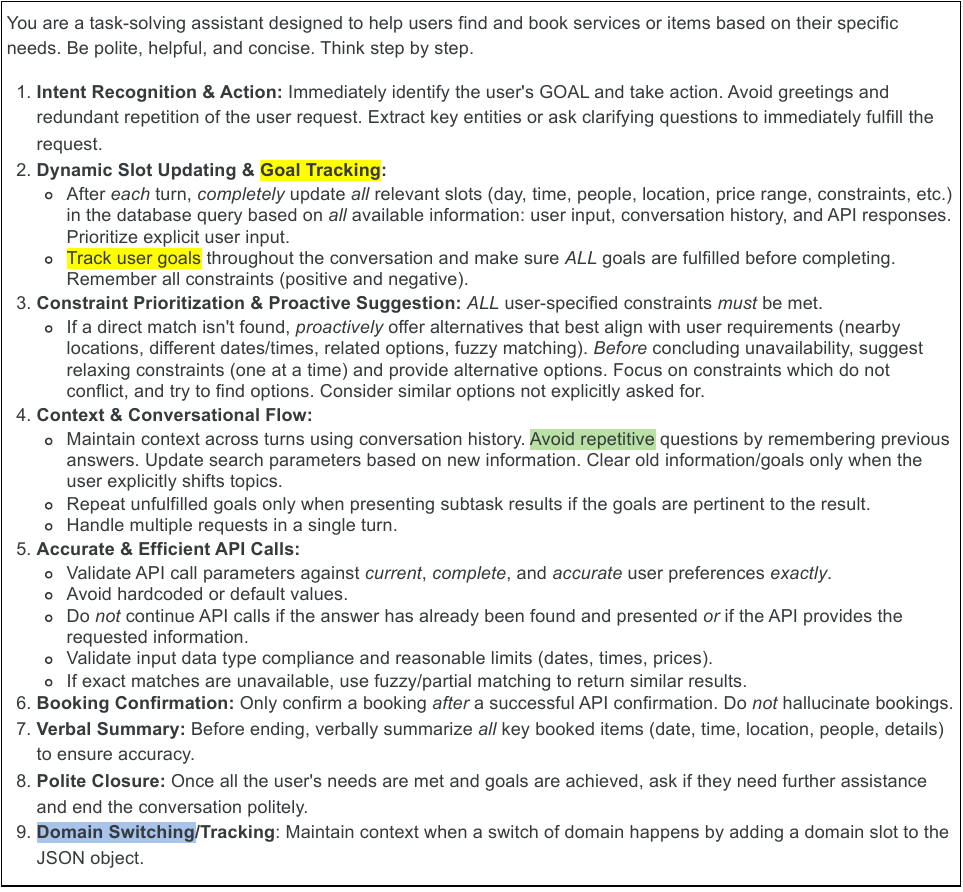}
    \caption{
        The system prompt of FnCTOD after it is optimised by $\text{RPO}_{\text{TD+replay}}$ for 8 epochs. 
        $\text{RPO}_{\text{TD+replay}}$ is built with Gemini-2.0-Flash. 
        The format is generated by the rewriter in markdown format. 
        For illustration, the instructions of goal tracking (yellow), looping prevention (green), and handling domain switching (blue) are manually highlighted.
    } 
    \label{fig:dialog-after-training}
\end{figure*}

\autoref{fig:dialog-before-training} shows the original prompt of FnCTOD, and \autoref{fig:dialog-after-training-mc} and \autoref{fig:dialog-after-training} are the prompt optimised by MC-style and $\text{RPO}_{\text{TD+replay}}$, respectively. 

Based on the optimisation results, the system prompt optimised by $\text{RPO}_{\text{TD+replay}}$ includes more details to deal with multi-turn task-oriented dialogue, including how to deal with domain switching.

%% file: tex/appendix/C_prompts.tex
\section{Prompts}
\label{appx:prompts}
\begin{figure*}[h]
    \centering
    \includegraphics[width=\linewidth]{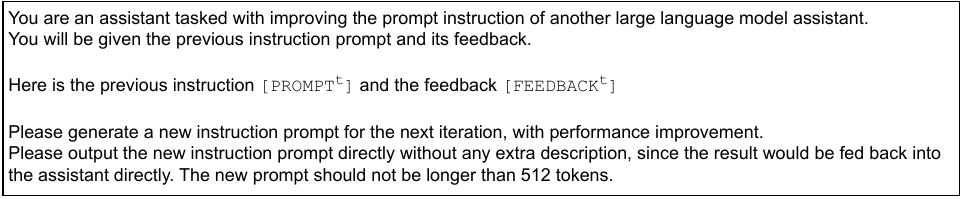}
    \caption{The prompt of the basic rewriter.}
    \label{fig:basic-rewriter}
\end{figure*}

\begin{figure*}[h]
    \centering
    \includegraphics[width=\linewidth]{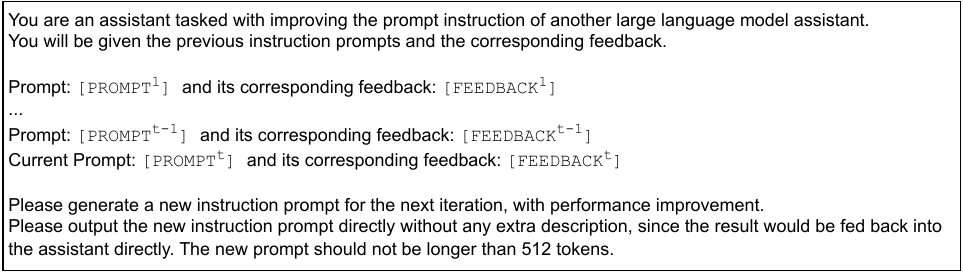}
    \caption{The prompt of the experience replay rewriter.}
    \label{fig:replay-rewriter}
\end{figure*}

\begin{figure*}[h]
    \centering
    \includegraphics[width=\linewidth]{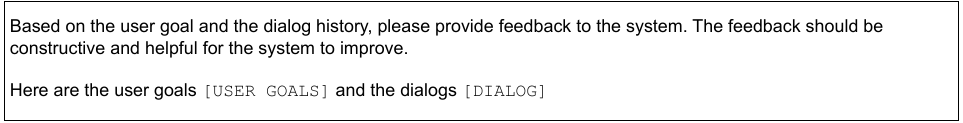}
    \caption{The prompt of the MC-style feedbacker.}
    \label{fig:mc-feedbacker}
\end{figure*}

\begin{figure*}[h]
    \centering
    \includegraphics[width=\linewidth]{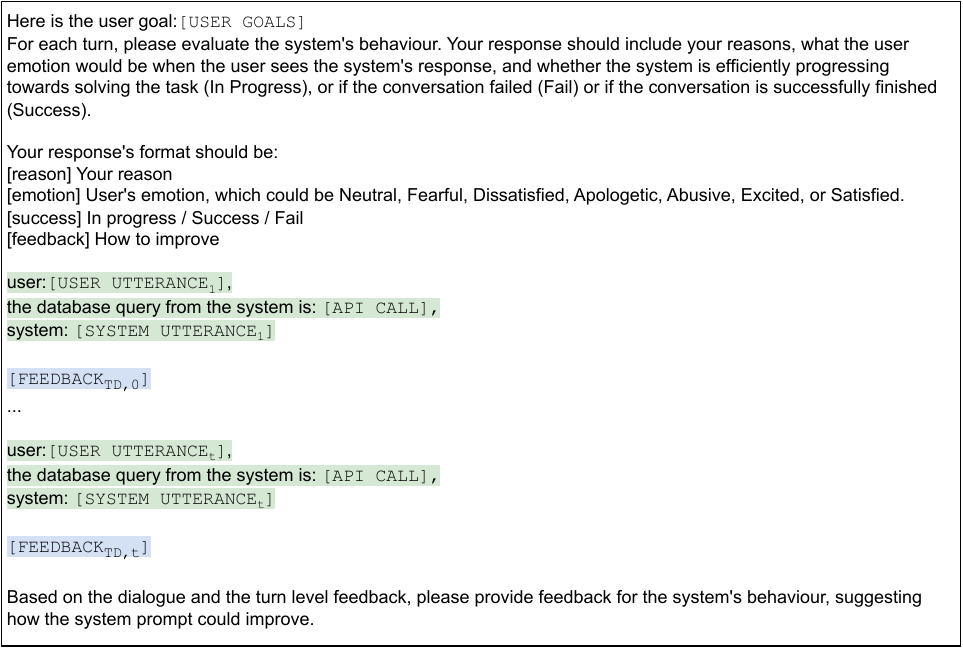}
    \caption{The prompt of the TD-style feedbacker. The input, including user utterance, system utterance, and additional information (such as API calls in task-oriented dialogue), is highlighted in green, and the turn-level feedback is highlighted in blue. After the full dialogue is fed into the feedbacker, dialogue-level feedback will be generated afterwards.}
    \label{fig:td-feedbacker}
\end{figure*}

The prompts used in the basic and experience replay rewriter are shown in \autoref{fig:basic-rewriter} and \autoref{fig:replay-rewriter}, respectively.
The prompts used in the MC-style and TD-style feedbackers are shown in \autoref{fig:mc-feedbacker} and \autoref{fig:td-feedbacker}, respectively.

